\documentclass[10pt,journal,compsoc]{IEEEtran}

\ifCLASSINFOpdf
\usepackage[pdftex]{graphicx}
\graphicspath{{./figures/}}

\else

\fi

\ifCLASSOPTIONcompsoc
\usepackage[nocompress]{cite}
\else
\usepackage{cite}
\fi

\usepackage[linesnumbered,lined,noend,ruled]{algorithm2e}
\usepackage{amsmath,amssymb}
\usepackage{subfigure}
\usepackage{amstext}
\usepackage{amsfonts}
\usepackage{url}
\usepackage{bbm}
\usepackage{color}

\usepackage{tablefootnote}
\usepackage{float}

\usepackage{color}
\definecolor{red}{RGB}{255,0,0}

\usepackage{multirow}
\usepackage[export]{adjustbox}
\usepackage{ragged2e}
\usepackage{booktabs}

\usepackage{pifont}       
\usepackage{bbding}       
\usepackage{fontawesome}  
 
\usepackage{xspace}

\newcommand{\eg}{\textit{e.g.}\xspace}

\begin{document}
\title{Video-Language Alignment via Spatio–Temporal Graph Transformer}
%
%

\author{Shi-Xue Zhang$ ^{\textbf{*}} $, Hongfa Wang$ ^{\textbf{*}} $, Xiaobin Zhu$ ^{\dagger} $, Weibo Gu, Tianjin Zhang, Chun Yang,\\ Wei Liu~\IEEEmembership{, IEEE Fellow}, Xu-Cheng Yin$ ^{\dagger} $~\IEEEmembership{, IEEE Senior Member}
\IEEEcompsocitemizethanks{
\IEEEcompsocthanksitem Shi-Xue Zhang, Xiaobin Zhu and Chun Yang are with the School of Computer and Communication Engineering, University of Science and Technology Beijing (USTB), Beijing, 100083, China. (E-mail: zhangshixue111@163.com; zhuxiaobin@ustb.edu.cn; chunyang@ustb.edu.cn.)
\IEEEcompsocthanksitem Xu-Cheng Yin is with the School of Computer and Communication Engineering, and Institute of Artificial Intelligence, University of Science and Technology Beijing (USTB), Beijing, 100083, China, also with USTB-EEasyTech Joint Lab of Artificial Intelligence, Beijing, 100083, China.(E-mail: xuchengyin@ustb.edu.cn.)
    
\IEEEcompsocthanksitem Weibo Gu, Tianjin Zhang, and Wei Liu are with the Tencent Hunyuan, Shenzhen, China. (e-mail: {lukegu, tildajzhang, vincentwliu}@tencent.com)

\IEEEcompsocthanksitem Hongfa Wang is with the Tencent, Hunyuan and Tsinghua Shenzhen International Graduate School. (e-mail: {hongfawang}@tencent.com)

\IEEEcompsocthanksitem ${\dagger}$ Corresponding authors ; $\textbf{*}$Authors contributed equally.
}
}

\markboth{Journal of \LaTeX\ Class Files,~Vol.~14, No.~8, May~2020}%
{Shell \MakeLowercase{\textit{et al.}}: Bare Demo of IEEEtran.cls for Computer Society Journals}

\IEEEtitleabstractindextext{%

\begin{abstract}
\justifying
Video-language alignment is a crucial multi-modal task for various downstream applications,\eg
, video-text retrieval and video question answering. Existing methods either utilize multi-modal information in video-text pairs or apply global and local alignment techniques to promote alignment precision. However, these methods often fail to fully explore the spatio-temporal relationships among vision tokens within video and across different video-text pairs. In this paper, we propose a novel \textbf{S}patio–\textbf{T}emporal \textbf{G}raph \textbf{T}ransformer module to uniformly learn spatial and temporal contexts for video-language alignment (dubbed \textbf{STGT}). Specifically, our STGT combines spatio-temporal graph structure information with attention in the transformer block to explore the spatio-temporal contexts fully. This way, we can effectively model the relationships between vision tokens to promote video-text alignment precision, greatly benefiting downstream tasks. In addition, we propose a novel cross-similarity alignment loss (\textbf{CSAL}) by evaluating the corresponding two video-video and text-text pairs to explore the inherent similarity after the initial optimization achieved by contrastive learning, which can further promote the aligning accuracy. Experimental results on challenging downstream tasks, including video-text retrieval and 
video question answering, verify the superior performance of our method. The code is available at: \url{https://github.com/GXYM/STGT}.

	
\end{abstract}


}

\maketitle

\IEEEdisplaynontitleabstractindextext

%
\IEEEpeerreviewmaketitle

\section{Introduction}\label{sec:introduction}
\IEEEPARstart{R}ecently, image-language pre-training has achieved significant success in cross-modal representation learning~\cite{blip,girdhar2023imagebind, zhu2023languagebind}. Adapting a robust image-language pre-trained model for video-language pre-training can offer significant advantages by leveraging knowledge derived from images. Some methods~\cite{videoclip, clip4clip,Vita-CLIP} directly applied image-language pre-trained models to video-text tasks, still outperforming video-specific models. As the critical task in vision-language pre-trained models, the power of video-language alignment can greatly determine the performance of various downstream tasks, such as video-text retrieval~\cite{videoclip,X-CLIP} and video question answering~\cite{Bridge_to_Answer,liu2021hair}.

\begin{figure}[htp]
    \setlength{\abovecaptionskip}{-0.10em}
	\begin{minipage}[t]{0.99\linewidth}
		\includegraphics[width=1\linewidth]{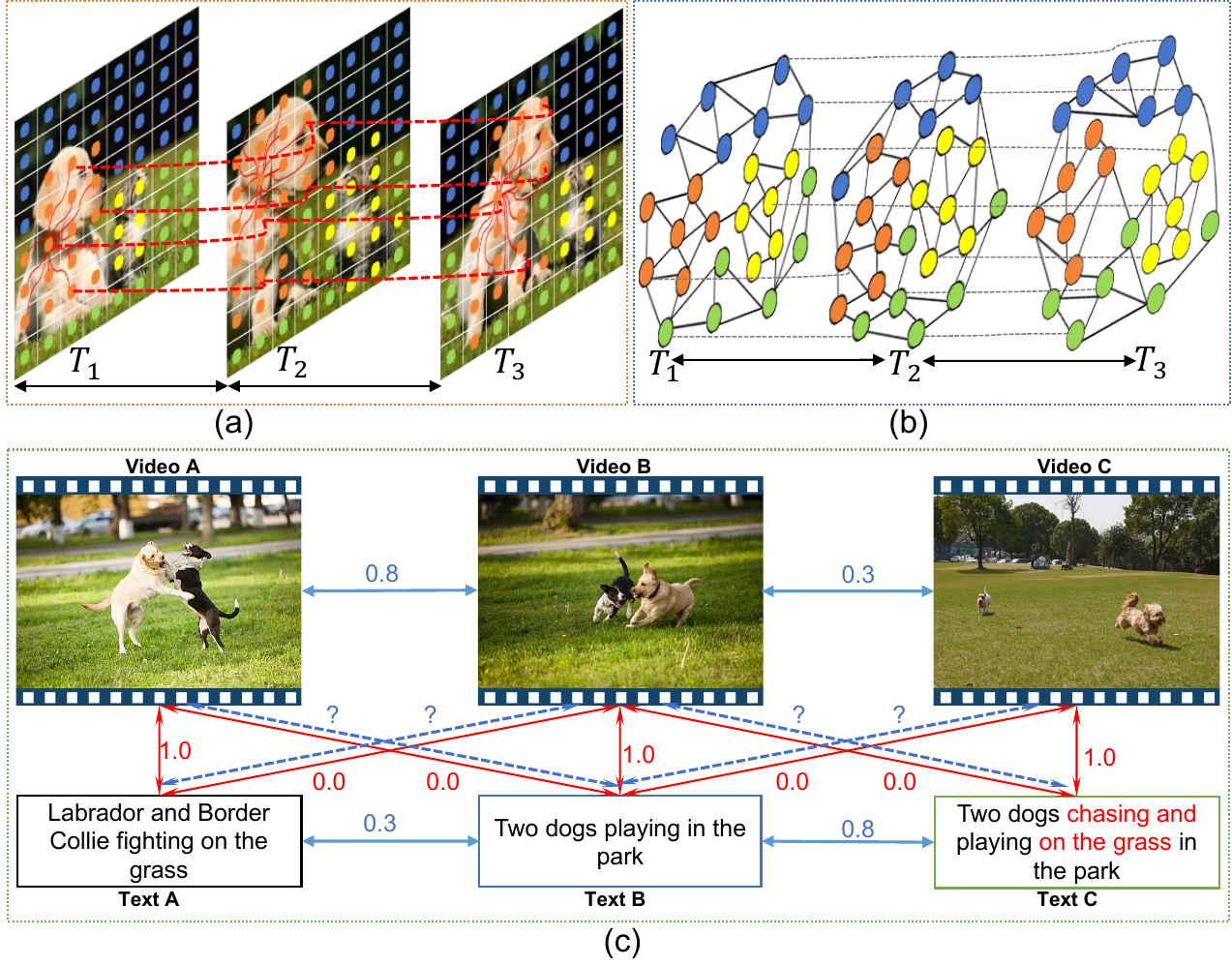}
		\caption{(a) The similarity between vision tokens in video. (b) The spatio-temporal graph. (c) The pair and similarity  relationship between video and texts.} \label{fig:intro1}
	\end{minipage}%
	\vspace{-0.5em}
\end{figure}

Some methods~\cite{clip4clip,Vita-CLIP} rely on pre-trained image-language models for video-language alignment training. CLIP4Clip~\cite{clip4clip} transfers knowledge from pre-trained CLIP~\cite{CLIP} for video clip retrieval. It adopts mean-pooling to compute similarity for averaging multiple frame features extracted from the pre-trained CLIP model and re-trains on the large-scale video-language Howto100M~\cite{miech2019howto100m} dataset. However, only averaging or fusing video frame features may not sufficiently capture the temporal dynamics and contextual information in videos. X-CLIP~\cite{X-CLIP} solves this problem by introducing a cross-frame communication transformer (CCT) and a multi-frame integration transformer (MIT). The CCT uses frame-level features from a pre-trained image-language model to capture semantics and model dependencies between frames, while MIT transfers frame-level features to video-level seamlessly. However, these strategies have yet to be fully validated in tasks, such as video-text retrieval and video question answering. Although some approaches have tried to capture temporal dynamics and inter-frame dependencies, exploring spatio-temporal relationships between vision tokens to boost feature representation capability remains to be improved.


Another problem is that the similarity relationship between different video-text pairs still needs to be further explored. Traditional contrastive learning based methods~\cite{X-CLIP, huang2023clover, videoclip} mainly focus on the matching relationship within annotated video-text pairs. In these methods, video-text pair A is deemed a positive sample, while pair B is considered a negative sample for A, as shown in Fig.~\ref{fig:intro1} (c). The contrastive loss aims to maximize the similarity between the embeddings of video A and text A while minimizing the similarity between the embeddings of video A and text B, as indicated by the red line in Fig.~\ref{fig:intro1} (c). However, it does not explicitly consider relationships between different videos or texts. In Fig.~\ref{fig:intro1} (c), videos A and B have similar scenes and content (similarity score is 0.8). However, texts A and B exhibit a lower similarity of 0.3 due to the differences in observation perspectives and language expression. Despite the low similarity between text A and B, text A can still describe video B and vice versa due to their high visual similarity. In such cases, it's unnecessary to force the similarity of learned embeddings between video A and text B, or learned embeddings between video B and text A to approach zero. Therefore, designing an objective loss that appropriately assigns similarity relationships to these cross video-text pairs is valuable for video-language alignment.

To address these problems, we propose a novel spatio–temporal graph transformer network for video-language alignment. Specifically, the spatio–temporal graph transformer module  combines the graph and transformer models to effectively and uniformly learn videos' spatial and temporal features. By measuring the similarity between vision tokens in video frames, we construct a spatio-temporal graph containing both temporal and spatial contextual information, as shown in Fig.~\ref{fig:intro1} (a). Then, the STGT directly integrates the spatio-temporal graph's topology and edge weights into the attention of the transformer block, effectively leveraging the spatio-temporal context offered by the graph structure. In this way, our method can model the relationships between vision tokens, promoting the precision of video-text alignment for benefiting downstream tasks. Additionally, we propose a novel cross-similarity alignment loss by evaluating the corresponding two video-video and text-text pairs to explore the inherent similarity after the initial optimization achieved by contrastive learning, which can further promote the aligning accuracy.  Experimental results on challenging downstream tasks, including video-text retrieval
and video question answering, verify the superior performance of our method. In summary, the main contributions of this paper are three-fold:

\vspace{-0.6em}
\begin{itemize}
\item We propose a novel spatio–temporal graph transformer module (\textbf{STGT}) that combines spatio-temporal graph structure information to fully explore spatial-temporal contexts between video and text pairs for video-language alignment.
\item We propose a novel cross-similarity alignment loss (\textbf{CSAL}) to explore  the inherent self-similarity via evaluating the corresponding two video-video and text-text pairs, further promoting the accuracy of video-text alignment. 
\item Experiments on challenging downstream tasks, \eg text-video retrieval and video question answering, verify the superior performance of our method.
\end{itemize}

\section{Related Work} \label{Related_Work}

\subsection{Vision-Language Pre-training}
Vision-language pre-training aims to develop a unified multi-modal representation, enhancing performance on various downstream tasks, such as video-text retrieval and video question answering. The dual-encoder is a common approach in visual-language alignment. Methods~\cite{Frozen, CLIP,clip4clip,ALIGN,videoclip,vatt,X-CLIP,zhang2023graph, zhou2023learning, fang2024transformer} employ two separate encoders to independently extract features for visual and textual data. CLIP~\cite{CLIP} effectively applies contrastive learning to learn image-language alignment from a large volume of noisy image-text pairs, achieving remarkable performance on vision-language tasks, as demonstrated in~\cite{wang2022cris,gao2022calic,liu2023vlpd, zhang2024inverse, zhang2021adaptive}. In VATT~\cite{vatt}, the authors employ contrastive learning to align the videos, audios and texts, and achieve impressive performance on the downstream tasks. Vita-CLIP~\cite{Vita-CLIP} introduces a multi-modal prompt learning scheme that balances supervised and zero-shot performance under a single unified training framework. 

Cross-fusion based methods~\cite{tan2019lxmert,li2020oscar, ALBEF, TCL,ViLT,xu2023bridgetower} use a cross-modal fusion encoder to enhance the interactions between vision and text features. BEiT-3~\cite{Beitv3} and VLMo~\cite{VLMo} employ modality-specific feed-forward networks and a shared self-attention layer in each block to facilitate flexible single-modality learning and cross-modal interactions. ALBEF~\cite{ALBEF} incorporates the image and text features into a cross-modal attention-based encoder to generate fused features. In ALPRO~\cite{ALPRO}, the authors introduce a video-text contrast (VTC) loss to align instance-level unimodal video-text features and design a prompt entity module to learn fine-grained alignment between visual regions and textual entities in a self-supervised manner. Following~\cite{blip}, BLIP-2~\cite{blip2} integrates pre-trained visual models with frozen parameters and large-scale language models.

\begin{figure*}[tp]
\setlength{\abovecaptionskip}{-0.00cm}
    \begin{center}
    \includegraphics[width=1.0\linewidth]{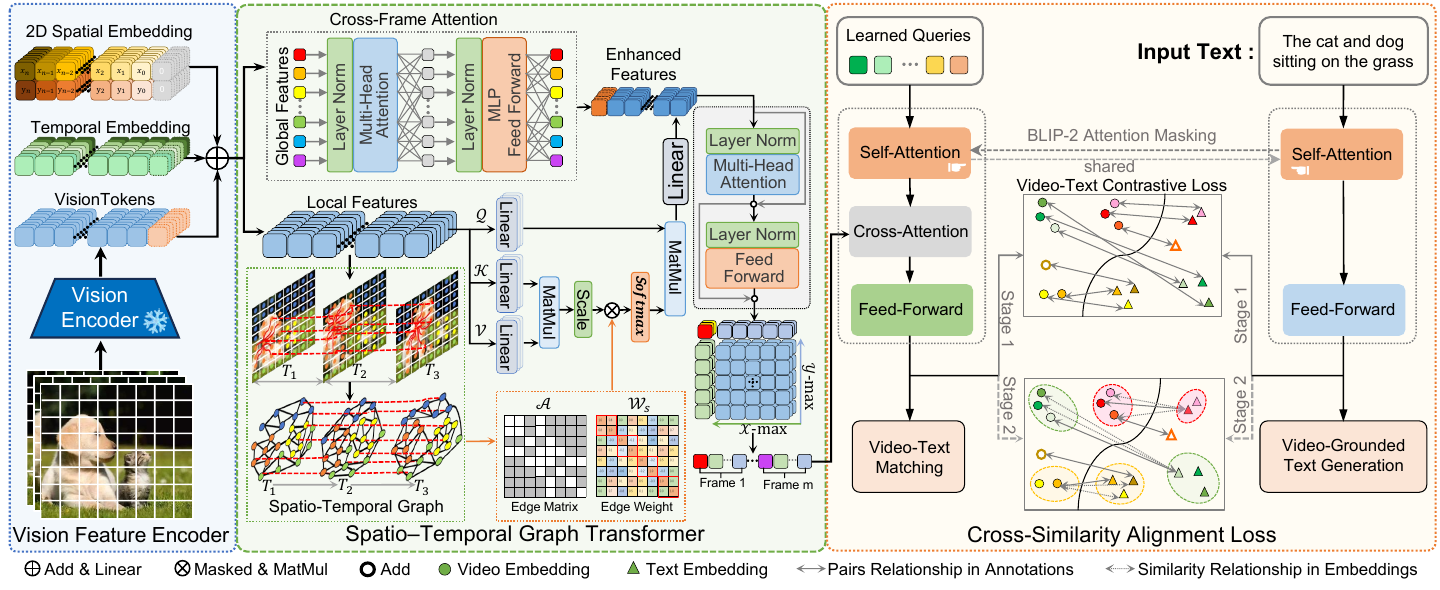 }
    \caption{Overview of the proposed framework. Firstly, vision tokens are obtained by a pre-trained ViT. Subsequently, a spatio-temporal graph transformer module is used to enhance global and local features. Finally, a self-similarity alignment loss is implemented to optimize video-text alignment following contrastive learning optimization. The BLIP-2~\cite{blip2} attention masking strategy for each objective to control query-text interaction.}
    \label{fig:framework}
    \end{center}
    \vspace{-1.5em}
\end{figure*} 

\vspace{-1.0em}
\subsection{Downstream Tasks}
\textbf{Video-Text Retrieval.} Previous research~\cite{yu2017end,yu2018joint} in video-text retrieval has primarily focused on developing intricate fusion mechanisms to facilitate cross-modal learning. Recently, vision-language pre-training models~\cite{videoclip,X-CLIP,clip4clip,T2vlad,wang2022align,han2021fine, zhang2022kernel, zhang2022arbitrary, zhang2023arbitrary} have made significant strides in zero-shot and fine-tuned video-text retrieval tasks. Clip4Clip~\cite{clip4clip} introduces a similarity calculator to directly transfer the robust knowledge from the pre-trained CLIP and continue pre-training on a large-scale video-language dataset for video clip retrieval. VideoClip~\cite{videoclip} trains a transformer for video and text by contrasting temporally overlapping positive video-text pairs with hard negatives from nearest neighbor retrieval. In DRL~\cite{DRL}, considering the inherent sequential structure in both text and video inputs, a weighted token-wise interaction module is implemented to decouple the content and exploit the pair-wise correlations. In~\cite{Li_2023_ICCV}, the authors introduce a text-video learning framework with progressive spatio-temporal prototype matching, which uncover semantic diversity in videos for dynamic matching.

\noindent\textbf{Video Question Answering.} Video question answering involves automatically generating responses to questions based on video context. This task necessitates a thorough understanding of both the video content and the language used in the questions. To extract more reliable multi-modal representations for the video question answering task, some methods~\cite{HSTT,PSAM,ahmad2023mmtf,li2023transformer,li2023redundancy} concentrate on enhancing spatio-temporal attention networks, while others~\cite{Bridge_to_Answer,liu2021hair,peng2023hierarchical} focus on designing superior question-video relation networks. Recently, the multi-modal encoder with video-language pre-training~\cite{chen2023tem,huang2023clover} has been employed to perform token-level cross-modal fusion. Clover~\cite{huang2023clover} uses a tri-modal alignment pre-training task to improve video-language feature alignment, and the pre-trained model is then applied to the video question answering task. VideoChat~\cite{li2023videochat} designs an end-to-end chat-centric video understanding system by pre-training a vision transformer (ViT) and a large language model (LLM).

\section{Method} \label{Proposed_Method}

\subsection{Overview}
We use a pre-trained vision transformer (ViT) as the vision encoder and a pre-trained BERT from BLIP-2~\cite{blip2} as the text encoder. As shown in Fig.~\ref{fig:framework}, our method consists of three main components. Firstly, we extract vision tokens using ViT and add  temporal and 2D spatial embeddings to enhance spatio-temporal relationships. Then, a spatio-temporal graph transformer module refines global and local features. This module constructs a spatio-temporal graph based on local token similarity and integrates the graph's topology and edge weights into attention for leveraging spatio-temporal context effectively. Additionally, a cross-frame attention block enhances cross-frame relationship of global information. Finally, we propose a cross-similarity alignment loss to optimize video-language alignment, supplementing contrastive loss for improving performance.

\vspace{-0.6em}
\subsection{Vision Feature Encoder}\label{Feature_Encoder}
For each video, we uniformly select $m$ frames as key-frames and employ a pre-trained vision transformer network (ViT)~\cite{CLIP} to extract visual features. Within the ViT, a video frame will be partitioned into fixed-size ($n \times n $) patches, which are then linearly embedded into a sequence of vector representations, referred to as vision tokens. These tokens contains local visual information, making them well-suited for processing within the transformer architecture. Additionally, a learnable $\mathbf{[CLS]}$ token is incorporated to extract the global features of the image.

In ViT, only one-dimensional position encoding 
$\{1, ...i, ..., n^2\}$ is introduced for each vision token. However, this is insufficient for understanding videos with 3D-dimension. To enhance the understanding of spatial and temporal relationships among these vision tokens, we introduce a 2D spatial and temporal embedding. The 2D spatial embedding transforms spatial coordinates into a high-dimensional space, preserving the relative or absolute positions of image patches. Typically, the $x$-coordinates $\{x_1,...,x_i,...,x_n\}$ and $y$-coordinates $\{y_1,...,y_i,...,y_n\}$ are transformed into high-dimensional spaces using sine and cosine functions, as
\begin{gather}
PE_{_{2i}}(z)=\sin(\dfrac{z}{10000^{{2i}/{d}}}), i \in  (0,d/2 ] \label{embed1}, \\ PE_{_{2i+1}}(z)=\cos(\dfrac{z}{10000^{{2i-1}/{d}}}), i \in  (0,d/2], \label{embed2}
\end{gather}
where $z$ is a scalar representing either $x$-coordinates or $y$-coordinates, and the dimension of the spatial embedding vector 
$PE(z)$ is $ d $. By applying the transformation 
$PE(z) $ to both $x$-coordinates and $y$-coordinates, we can derive a 2D spatial embedding. Each vision token is then augmented with a 2D spatial embedding, which indicates its original position within the image grid.

Besides spatial encoding, we apply a set of learnable temporal embedding $\{T_1,...,T_i,...,T_m\}$ to vision tokens. This embedding contains temporal information, which allows the model to encode contextual information over time, thereby understanding the temporal dynamics in video. Both embeddings (2D spatial embedding $E_{2p}$ and temporal embedding $E_{t}$) are then combined with vision tokens by element-wise addition, denoted as:
\begin{equation}
F_v(j) = [Tk_{{cls}}(j), Tk_1(j),...,Tk_{n^2}(j)]+ E_{2p}+E_{t},
\end{equation}
where $Tk_{i}(j)$ represents the i-$th$ vision token of the j-$th$ frame in video; $F_v(j)$ represents the features of the j-$th$ frame in video.  The range of $i$ is from $1$ to $n^2$, and the range of $j$ is from $1$ to $m$. The combined embedding $F_v(j)$ incorporates the vision tokens, 2D spatial embedding $E_{2p}$, and temporal embedding $E_{t}$. This strategy allows the spatio-temporal graph transformer module to access both raw visual data and spatial-temporal context. By linking different tokens, the model learns a context-rich representation of each token, thereby enhancing its understanding of the video sequence.

\begin{figure}[tp]
\setlength{\abovecaptionskip}{-0.05em}
	\begin{minipage}[t]{0.95\linewidth}
		\includegraphics[width=1\linewidth]{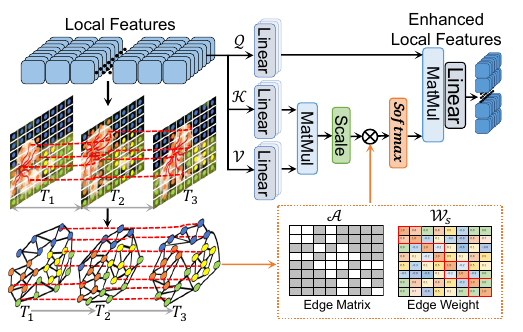}
		\caption{Illustrations of the local spatio-temporal graph transformer.} \label{fig:local}
	\end{minipage}%
	\vspace{-0.8em}
\end{figure}

\subsection{Spatio–Temporal Graph Transformer}

\textbf{Global Features.} The $\mathbf{[CLS]} $ vision token is important as it holds the global features ($\mathcal{X}_g$) of each video frame. These features give an important global representation of the video frame. In Section~\ref{Feature_Encoder}, we use temporal embeddings to encode the time position of the frame in a video sequence. By merging these with global features, our method considers the video's temporal attributes. To enable the exchange of global information across frames, we employ a cross-frame attention block, as
\vspace{-0.1em}
\begin{gather}
{\mathcal{X}_g}^{'} = MSA(LN(\mathcal{X}_g))+\mathcal{X}_g, \\{\mathcal{X}_g}^{''}= MLP(LN({\mathcal{X}_g}^{'}))+{\mathcal{X}_g}^{'},
\end{gather}
where $MSA$ represents multi-head attention,  $LN$ stands for layer normalization, and $MLP$ denotes multi-layer perceptron. The cross-frame attention captures temporal dynamics and inter-dependencies between frames, thereby enriching the model's comprehension of the video content.

\textbf{Local Features.} The local features ($\mathcal{X}_l$) are primarily contained within local vision tokens ($n \times n$), excluding the $\mathbf{[CLS]} $ vision token. These tokens, derived from a pre-trained vision transformer (ViT) model, include detailed information about specific regions in  frame. Generally, local tokens from the same object region give higher similarity than those from different regions. This observation underpins various methods that employ CLIP~\cite{CLIP} for downstream tasks such as instance segmentation (CLIPSeg~\cite{clipseg}), semantic segmentation (DenseCLIP~\cite{Denseclip}), and object detection (RegionCLIP~\cite{Regionclip}, VLPD~\cite{liu2023vlpd}).

In videos, the observed similarity principle extends beyond individual frames to include multiple frames. By effectively utilizing  these similarities, local features can be thoroughly enhanced. In our method, we construct a spatio-temporal graph ($g(\mathcal{X}_l, \mathcal{A})$) based on the similarities of local tokens. Each local vision token is regarded as a node in the graph. Consequently, the local feature matrix can directly represent the node feature matrix ($\mathcal{X}_l=flatten([F_v(1),...,F_v(i),...,F_v(m)],(0, 1))$) of the constructed graph. These features integrate both the positional and temporal information of the tokens, enabling comprehensive spatio-temporal reasoning within the graph. The topology of the spatio-temporal graph can be represented by adjacency matrix $\mathcal{A}$. The adjacency matrix $\mathcal{A}$ defines the connections between nodes in the graph, and contains the spatio-temporal relationships between tokens across different video frames. 

\textbf{Spatio-Temporal Graph.} To capture both spatial and temporal relationships in a video, we compute the similarity among all tokens within each frame, thereby constructing a spatial graph that represents the spatial relationships among local vision tokens. In the temporal dimension, we focus on the similarity between vision tokens in consecutive frames, capturing temporal connections via a temporal graph. This strategy effectively models the temporal relationships and dependencies within the video. To integrate the spatial and temporal context information, we merge the spatial and temporal graphs into a spatio-temporal graph by linking joint nodes between the two graphs, as shown in Fig.~\ref{fig:local}. The spatio-temporal graph offers a comprehensive representation of relationships between local vision tokens.

\begin{figure}[tp]
\setlength{\abovecaptionskip}{-0.00em}
	\begin{minipage}[t]{0.99\linewidth}
		\includegraphics[width=1\linewidth]{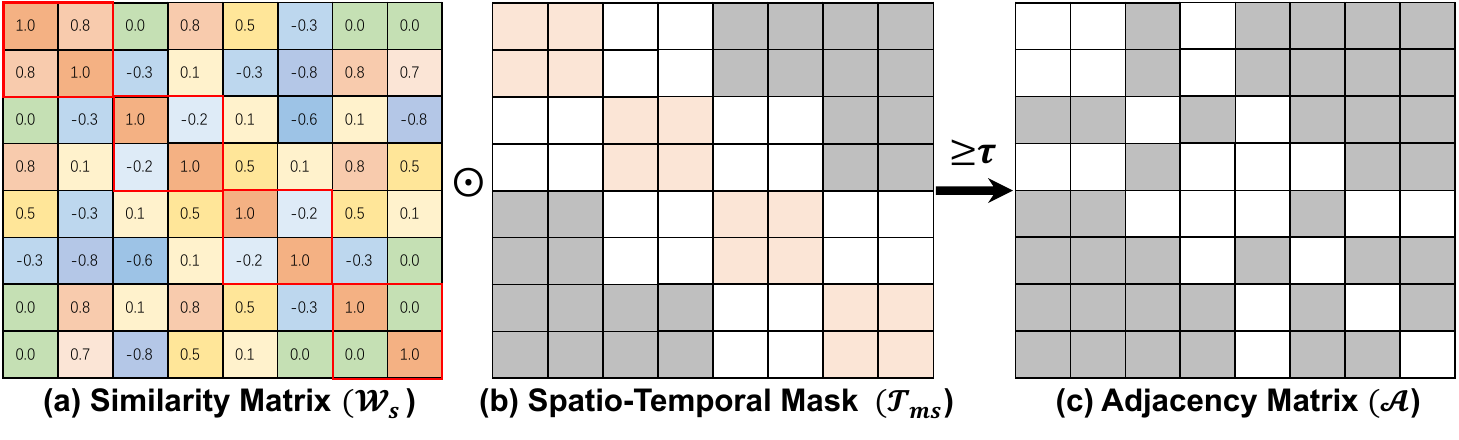}
	\caption{Illustration of adjacency matrix generation. (a) Similarity matrix $\mathcal{W}_s$; (b) Spatio-temporal mask  $\mathcal{T}_{mask}$; (c) Adjacency matrix $\mathcal{A}$. Each frame is depicted with two blocks. In (b), yellow denotes intra-frame similarity, white denotes inter-frame similarity, and gray denotes masked similarity. }
 \label{fig:matrix}
	\end{minipage}%
	\vspace{-0.8em}
\end{figure}
Since the similarities between local tokens within frames and between adjacent frames are easily obtained, we can use this information to determine the connections in the spatio-temporal graph. To get the adjacency matrix $\mathcal{A}$, we can iterate over all the nodes in the graph and compute their similarities, as
\begin{gather}
\mathcal{W}_s = \mathcal{X}_l \cdot {\mathcal{X}_l}^ \mathrm{ T }, \\
\mathcal{A} = \begin{cases} \; 1, \;\; & \text{if } \;\; \mathcal{W}s \odot \mathcal{T}_{ms} \geq \hbar, 
\\ 
\;0, \;\; & \text{if } \;\; \mathcal{W}s \odot \mathcal{T}_{ms} < \hbar, \end{cases}, \label{EQ_A}
 \end{gather}
where $\mathcal{X}_l$ represents the node feature matrix with dimensions ($n \times n \times m$, $d$); $\mathcal{W}_s $ denotes the similarity matrix of all nodes in the spatio-temporal graph; $\mathcal{T}_{ms}$ is a mask used to ensure that edges are only established within local tokens of each frame or between adjacent frames in the video. If the similarity between two tokens exceeds a threshold ($\hbar$), they are deemed as connected nodes in the graph, and the corresponding value in $\mathcal{A}$ is assigned a weight of 1, as shown in Fig.~\ref{fig:matrix}. During training, $\hbar$ is set to 0.1 by default. By Eq.~\ref{EQ_A}, we can efficiently generate the adjacency matrix of the spatio-temporal graph, capturing the linkages and edge weights between nodes based on their similarities.

\begin{figure}[tp]
\setlength{\abovecaptionskip}{-0.00em}
	\begin{minipage}[t]{0.95\linewidth}
		\includegraphics[width=1\linewidth]{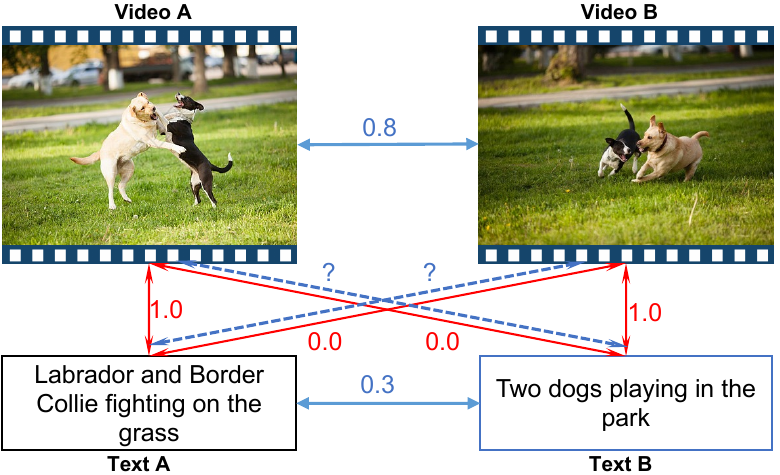}
		\caption{ Illustrations of relationships in  videos and texts.} \label{fig:ssal-1}
	\end{minipage}%
	\vspace{-1.5em}
\end{figure}

\textbf{Graph Transformer.} Existing methods~\cite{HSTT,Bridge_to_Answer,zhang2020deep} often directly use graph networks in their architectures, requiring many additional operations for data format conversion. Inspired by the graph representation learning method~\cite{ying2021do}, we directly integrate the spatio-temporal graph's topology and edge weights into the attention mechanism for improving effectiveness, by the following formulas:
\vspace{-0.1em}
\begin{gather}
Q = \mathcal{X}_lW^Q, \;\;K = \mathcal{X}_lW^K, \;\;V = \mathcal{X}_lW^V,\\
A = \dfrac{QK^T}{\sqrt{d_K}}\odot\mathcal{A}\odot\mathcal{W}_s, \\
A_{ms} = \mathcal{A} \cdot 1 + (1 - \mathcal{A}) \cdot (-\infty),\label{EQ_Mask}\\
Attention(\mathcal{X}_l) = Softmax(A \odot A_{ms})V,\label{EQ_Atten}\\
{\mathcal{X}_l}^{''}= Attention(\mathcal{X}_l)W^l,
\end{gather}
where $d_K$ represents the dimension size of $K$, and $\odot$  denotes the element-wise product.
Applying Eq.~\ref{EQ_Mask} and Eq.~\ref{EQ_Atten}, our model can successfully incorporate the topological information and edge weights of the spatio-temporal graph into the transformer's attention mechanism. This integration improves the fusion of local features, enabling more effective learning and reasoning for the spatio-temporal graph. The proposed spatio–temporal graph transformer module
presents a simpler yet highly effective alternative of traditional graph networks, improving the ability of video feature learning and reasoning.

\textbf{Max-Pooling Sampling.} To facilitate the interaction of global and local information, the enhanced global and local features ($\mathcal{X}_g$ and $\mathcal{X}_l$) are concatenated and subsequently fed into a residual transformer block, as
\begin{gather}
X = {\mathcal{X}_g}^{''} || \; {\mathcal{X}_l}^{''}, \; {X}^{'} = MSA(LN(X))+X,\\
{X}^{''}= MLP(LN({X}^{'}))+{X}^{'},
\end{gather}
where $||$ indicates a concatenation operation. By residual transformer block, we effectively integrate both global and local information to capture the synergistic relationship between global and local contexts, thereby enhancing the comprehensive understanding of the video. To reduce computational complexity, we employ max-pooling to sample the local features, as
\begin{equation}
X = X_g'' \; || \; max(X_l'', d_x) \; || \; max(X_l'', d_y),
\end{equation}
where $X_g''$ and $X_l''$ represent global and local information in feature ${X}^{''}$, respectively; $d_x$ and $d_y$ indicate sampling along the $x$-dimensions and $y$-dimensions. Using max-pooling to down-sample local features retains important information while reducing computation, enabling efficient processing and faster inference in subsequent modules without sacrificing performance.

\vspace{-0.0em}
\subsection{Cross-similarity Alignment Loss}
In traditional contrastive learning, only the video-text pairs with the same annotation are considered as positive samples, while other video-text pairs are treated as negative samples. As shown in Fig.~\ref{fig:ssal-1}, the video-text pair A is regarded as a positive sample, while pair B is regarded as a negative sample for A. The contrastive loss aims to maximize the similarity between the
embeddings of video A and text A (toward one), while minimizing the similarity between the embeddings of video A and text B, as shown in Fig.~\ref{fig:ssal-1} with a red line.

Traditional contrastive learning focuses on the alignment of video and text within the same annotated pairs, but overlooks relationships across different pairs. In Fig.~\ref{fig:ssal-1}, videos A and B have similar scene and content ( similarity score is 0.8). However, texts A and B exhibit a lower similarity of 0.3 due to the differences in observation perspectives, language expression and content summarization. Despite the low similarity between text A and B, text A can still describe video B and vice versa, due to their high visual similarity. In such cases, it's unnecessary to force the similarity of learned embeddings between video A and text B, or learned embeddings between video B and text A to approach zero. For cross video-text pairs, new similarity relationships need to be considered, as indicated by the blue line in Fig.~\ref{fig:ssal-1}.

\begin{figure}[tp]
\setlength{\abovecaptionskip}{-0.00em}
	\begin{minipage}[t]{0.99\linewidth}
		\includegraphics[width=1\linewidth]{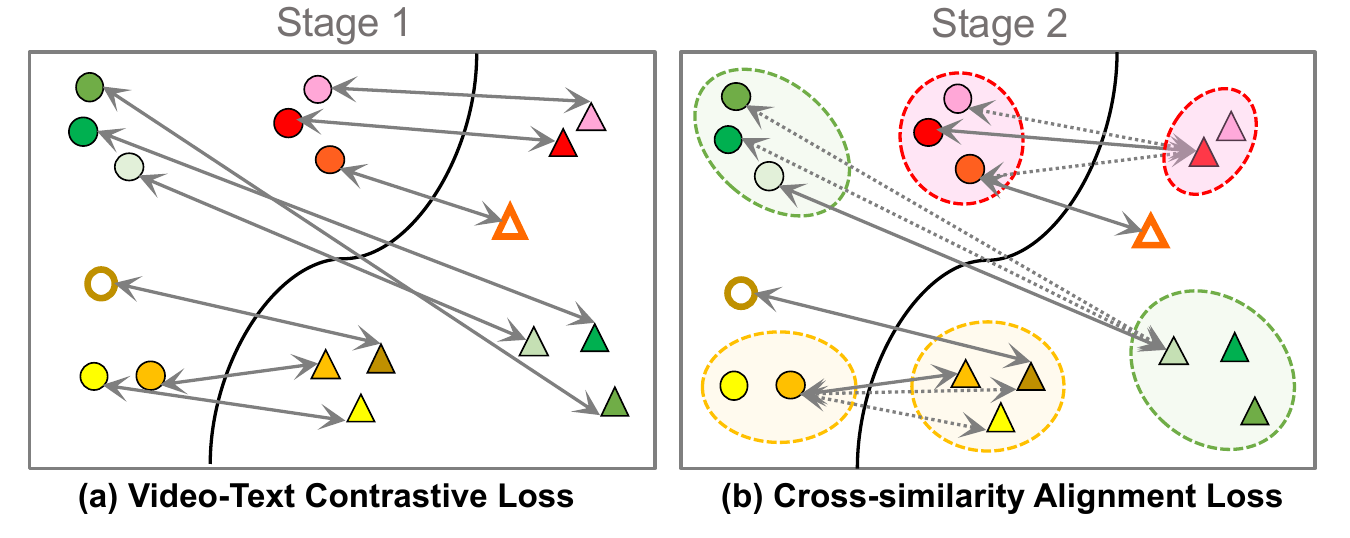}
		\caption{(a) Contrastive learning: optimizing relationships between annotated video-text pairs. (b) Self-similarity alignment: exploring cross-video text pair relationships through visual/textual similarities. Solid and dashed arrows represent annotated pairings and similarity relationships, respectively. Solid shapes indicate high similarity, while hollow shapes denote dissimilarity within the same modality.} \label{fig:ssal-2}
	\end{minipage}%
	\vspace{-1.5em}
\end{figure}

To explore the similarity relationships in cross video-text pairs for further enhancing video-text alignment, we propose a cross-similarity alignment loss with the the initial optimization by contrastive learning. In initial stage, we employ contrastive learning loss ($\mathcal{L}_{vtc}$) to guide the model's optimization, focusing on the point-to-point relationships between annotated video-text pairs. The $\mathcal{L}_{vtc}$ helps the model understand the basic video-language alignment, thereby facilitating convergence. The form of $\mathcal{L}_{vtc}$ is as follows:
\begin{gather}
P_{_{v2t}}^{(i)} = \frac{e^{S(v_i, t_i)/{\tau}}}{\sum_{k=1}^{B} e^{S(v_i, t_k)/{\tau}}}, \;\; P_{_{t2v}}^{(i)} = \frac{e^{S(t_i, v_i)/{\tau}}}{\sum_{k=1}^{B} e^{S(t_i, v_k)/{\tau}}},\\
\mathcal{L}_{vtc} =  -\frac{1}{2B}({\sum_{i=1}^{B} Y_{_{v2t}}^{(i)} log(P_{_{v2t}}^{(i)}}) + {\sum_{i=1}^{B} Y_{_{t2v}}^{(i)} log(P_{_{t2v}}^{(i)}})),
\end{gather}
where $B$ denotes the batch size, and $\tau$ serves as a learnable temperature parameter; $ S(v, t)$ and $ S(t, v)$ denote the similarity between video-text and text-video; $Y_{v2t}$ and $Y_{t2v}$ denote the ground-truth one-hot similarity where negative pairs have a probability of 0 and positive pair has a probability of 1.

In cross-similarity alignment stage, we adopt the cross-similarity alignment loss ($\mathcal{L}_{csal}$) to further refine video-language alignment by leveraging the inherent cross-similarity present in the video and text data. The formula of the proposed $\mathcal{L}_{csal}$ is as follows:
\begin{gather}
S_{_{v \cdot t}}^{(i,j)} = \begin{cases}
-\infty, & \text{if} \; S(v_i, v_j) \land S(t_i, t_j) \leq 0, \\
S(v_i, v_j) \cdot S(t_i, t_j), & \text{otherwise},
\end{cases}\\
P_{_{v2t}}^{(i,j)} = \frac{e^{S(v_i, t_j)/{\tau}}}{\sum_{k=1}^{B} e^{S(v_i, t_k)/{\tau}}}, \;\; P_{_{t2v}}^{(i,j)} = \frac{e^{S(t_i, v_j)/{\tau}}}{\sum_{k=1}^{B} e^{S(t_i, v_k)/{\tau}}},\\
\mathcal{L}_{csal} =  -\frac{1}{2B}{\sum_{i=1}^{B}\sum_{j=1}^{B}
\frac{e^{{\gamma} \cdot S_{_{v \cdot t}}^{(i,j)}
}}{\sum_{k=1}^{B} e^{{\gamma} \cdot S_{_{v \cdot t}}^{(i,k)}}}(log(P_{_{v2t}}^{(i,j)}}) + log(P_{_{t2v}}^{(i,j)})),
\end{gather}
where $S(v, v$ and $S(t, t)$ are the similarity between video-video and text-text, respectively. Unlike $\mathcal{L}_{vtc}$, the $\mathcal{L}_{csal}$ focuses on distribution-to-distribution relationships by leveraging the inherent cross-similarity within the video and text data, as shown in Fig.~\ref{fig:ssal-2}. With the $\mathcal{L}_{csal}$, the alignment accuracy and overall performance are both improved. The hyper-parameter $\gamma$ must be greater than zero.

In addition to video-text contrastive loss $\mathcal{L}_{vtc}$ and the cross-similarity alignment loss $\mathcal{L}_{csal}$, we also use the vision-text matching loss $\mathcal{L}_{vtm}$ and vision-grounded text generation loss $\mathcal{L}_{vtg}$, following BLIP-2~\cite{blip2}. Consequently, the total loss of our method can be expressed as follows:
\begin{gather}
\mathcal{L} = \alpha \cdot \mathcal{L}_{vtc} + (1-\alpha) \cdot \mathcal{L}_{csal} + \mathcal{L}_{vtm} + \mathcal{L}_{vtg}
\end{gather}
where parameter  $\alpha$ is set to one in the initial training stage and zero in cross-similarity alignment training stage.

\section{Experiments} \label{Experiments}
\subsection{Datasets and Downstream Tasks}

\textbf{Pre-training datasets.} Following recent work~\cite{chen2023tem,zou2023spaceclip,zou2023spaceclip,cheng2023vindlu}, 
we pre-train our STGT on Webvid2M~\cite{Frozen}, CC3M\cite{sharma2018conceptual}, Webvid-10M~\cite{Frozen} and VIDAL-10M~\cite{zhu2023languagebind} (we only obtain 9.6M video-text pairs in WebVid-10M and 7M video-text pairs in VIDAL-10M due to video url broken or file broken.)

\textbf{MSRVTT~\cite{xu2016msr}} contains 10K videos with 200K text captions. Follow~\cite{clip4clip}, we use 9k splits as train set. The test set is ‘test 1k’, which contains 1,000 clip text pairs.

\textbf{DiDeMo~\cite{anne2017localizing}} contains 10k videos with 40k text descriptions. We evaluate video-paragraph retrieval following~\cite{ALPRO}, where all text for a video are combined into a single query.

\textbf{LSMDC~\cite{rohrbach2017movie}} consists of 118,081 video clips  with one caption corresponding to each clip. Evaluation is conducted on a test set of 1,000 videos.

\textbf{MSVD~\cite{chen2011collecting}} contains 1,970 videos. Train, validation, test splits contain 1,200, 100, and 670 videos, respectively. Each video has about 40 associated sentences in English.

\textbf{MSVD-QA~\cite{ALPRO}} is built upon videos and text from MSVD. The MSVD-QA has a total 1,970 videos and 50k question answer pairs, with 2,423 answer candidates.

\textbf{MSRVTT-QA~\cite{ALPRO}} is built upon videos and captions from MSRVTT, which contains 10k videos with 243k  questions and 1.5k answer candidates.

\begin{figure}[tp]
\setlength{\abovecaptionskip}{-0.00em}
	\begin{minipage}[t]{0.99\linewidth}
		\includegraphics[width=1\linewidth]{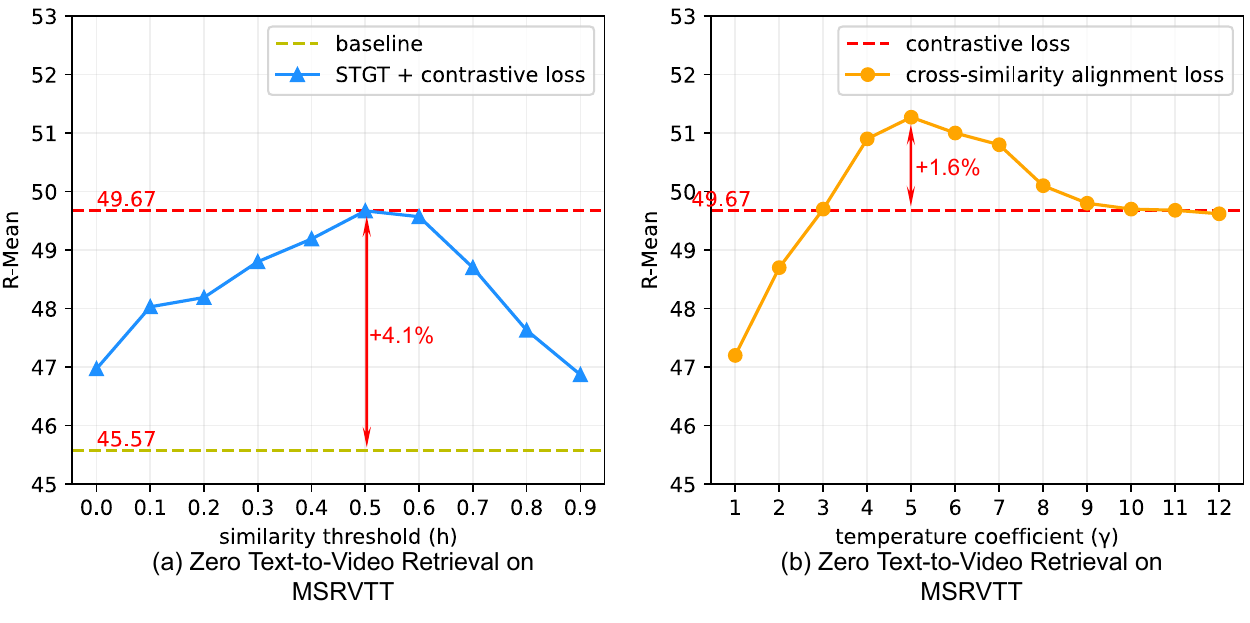}
		\caption{ Zero-shot results of text-to-video
retrieval on MSRVTT with Webvid-10M pre-training for similarity threshold $\hbar$ and temperature coefficient $\gamma$. The `baseline' means that we use one layer of cross frame communication transformer and multi-frame integration transformer from X-CLIP~\cite{X-CLIP} for video feature fusion and enhancement.} \label{fig:hyperparameter}
	\end{minipage}%
	\vspace{-1.5em}
\end{figure}

\begin{table*}[tp]
\centering
\caption{Text-to-video retrieval performance  comparison under {fine-tune} and {zero-shot} setups. Here {higher} R@k (Recall K) and {higher} R-mean (Mean Recall) indicate better performance. W2M, C3M, H100M, HDV100M, Y180M, are short for Webvid2M~\cite{Frozen}, CC3M\cite{sharma2018conceptual}, HowTo100M\cite{miech2019howto100m},
HD-VILA-100M\cite{xue2022advancing},
YT-Temporal-180M\cite{zellers2021merlot},  VIDAL-4M \& 7M~\cite{zhu2023languagebind}, respectively. `$\ddag$' indicates that our video and text embeddings use a 1024 dimension, with a default of 256.}
\vspace{-0.5em}
\renewcommand{\arraystretch}{0.65}
\resizebox{0.98\textwidth}{!}{
\begin{tabular}{cccccccccccccccccc}
\toprule
\multicolumn{1}{l|}{\multirow{2}{*}{Method}} & \multicolumn{1}{c|}{\multirow{2}{*}{Published}} &
\multicolumn{1}{c|}{\multirow{2}{*}{Pre-training dataset }}& \multicolumn{5}{c|}{DiDeMo} & \multicolumn{5}{c|}{MSRVTT} & \multicolumn{5}{c}{LSMDC} \\ \cline{4-18} 
\multicolumn{1}{l|}{} &\multicolumn{1}{l|}{} & \multicolumn{1}{c|}{} & R@1$\uparrow$ & R@5$\uparrow$ & R@10$\uparrow$ &R-Mean$\uparrow$ & \multicolumn{1}{c|}{MedR$\downarrow$} & R@1$\uparrow$ & R@5$\uparrow$ & R@10$\uparrow$ &R-Mean$\uparrow$ & \multicolumn{1}{c|}{MedR$\downarrow$} & R@1$\uparrow$ & R@5$\uparrow$ & R@10$\uparrow$ & R-Mean$\uparrow$ &MedR$\downarrow$\\ \midrule
\multicolumn{16}{c}{\textbf{Fine-tune}} \\ \midrule
\multicolumn{1}{l|}{Frozen \cite{Frozen}} & \multicolumn{1}{c|}{ICCV'21} & \multicolumn{1}{c|}{W2M+C3M} & 31.0 & 59.8 & 72.4 &54.40 & \multicolumn{1}{c|}{3} & 31.0 & 59.5 & 70.5 &53.67 & \multicolumn{1}{c|}{3} & 15.0 & 30.8 & 39.8 & 28.53 &20.0\\
\multicolumn{1}{l|}{HD-VILA \cite{xue2022advancing}} & \multicolumn{1}{c|}{CVPR'22} & \multicolumn{1}{c|}{HDV100M} & 28.8 & 57.4 & 69.1 &51.77 & \multicolumn{1}{c|}{4} & 35.6 & 65.3 & 78.0 &59.63 & \multicolumn{1}{c|}{3} & 17.4 & 34.1 & 44.1 & 31.87 &15.0\\

\multicolumn{1}{l|}{ALPRO \cite{ALPRO}} & \multicolumn{1}{c|}{CVPR'22} & \multicolumn{1}{c|}{W2M+C3M} & 35.9 & 67.5 & 78.8 &60.73 & \multicolumn{1}{c|}{3} & 33.9 & 60.7 & 73.2 &55.93 & \multicolumn{1}{c|}{3} & - & - & - & - &-\\
\multicolumn{1}{l|}{TMVM \cite{lin2022text}} & \multicolumn{1}{c|}{NeurIPS'22} & \multicolumn{1}{c|}{W2M+C3M} & 36.5 & 64.9 & 75.4 &58.93 & \multicolumn{1}{c|}{3} & 36.2 & 64.2 & 75.7 &58.70 & \multicolumn{1}{c|}{3} & 17.8 & 37.1 & 45.9 & 33.60 &13.5\\
\multicolumn{1}{l|}{OA-Trans \cite{wang2022object}} & \multicolumn{1}{c|}{CVPR'22} & \multicolumn{1}{c|}{W2M+C3M}& 34.8 & 64.4 & 75.1 & 58.10 & \multicolumn{1}{c|}{3}  & 35.8 & 63.4 & 76.5 &58.57 & \multicolumn{1}{c|}{3} & 18.2 & 34.3 & 43.7 & 32.07 &18.5\\
\multicolumn{1}{l|}{X-Pool~\cite{X-Pool}} & \multicolumn{1}{c|}{ CVPR'22} & \multicolumn{1}{c|}{-} & - & - & - &- & \multicolumn{1}{c|}{-} & 46.9 & 72.8 & 82.2 &67.30 & \multicolumn{1}{c|}{2} & 25.2 & 43.7 & 53.5 & 40.80 &8.0\\ 
\multicolumn{1}{l|}{CLIP4Clip~\cite{clip4clip}} & \multicolumn{1}{c|}{ NeurCom'22} & \multicolumn{1}{c|}{W10M+H100M} & 43.4 & 70.2 & 80.6 &64.73 & \multicolumn{1}{c|}{2} & 44.5 & 71.4 & 81.6 &65.83 & \multicolumn{1}{c|}{2} & 21.6 & 41.8 & 49.8 & 37.73 &11.0\\ 
\multicolumn{1}{l|}{TVTS~\cite{zeng2023learning}} & \multicolumn{1}{c|}{CVPR'23} & \multicolumn{1}{c|}{Y180M} & 32.4 & - & - &- & \multicolumn{1}{c|}{-} & 34.6 & - & - &- & \multicolumn{1}{c|}{-} & 17.2 & - & - & - &-\\ 
\multicolumn{1}{l|}{All-in-1 \cite{wang2023all}} & \multicolumn{1}{c|}{CVPR'23} & \multicolumn{1}{c|}{W2M+H100M} & 32.7 & 61.4 & 73.5 &55.87 & \multicolumn{1}{c|}{-} & 37.9 & 68.1 & 77.1 &61.03 & \multicolumn{1}{c|}{-} & - & - & - & - &-\\
\multicolumn{1}{l|}{ProST~\cite{Li_2023_ICCV}} & \multicolumn{1}{c|}{ICCV'23} & \multicolumn{1}{c|}{-} & 44.9 & 72.7 & 82.7 &66.77 & \multicolumn{1}{c|}{2} & 48.2 & 74.6 & 83.4 &68.73 & \multicolumn{1}{c|}{2} & 24.1 & 42.5 & 51.6 & 39.40 &9.0\\ 
\multicolumn{1}{l|}{Clover~\cite{huang2023clover}} & \multicolumn{1}{c|}{CVPR'23} & \multicolumn{1}{c|}{W2M+C3M} & 50.1 & 76.7 & 85.6 &70.80 & \multicolumn{1}{c|}{1} & 40.5 & 69.8 & 79.4 &63.23 & \multicolumn{1}{c|}{2} & 24.8 & 44.0 & 54.5 & 41.10 &8.0\\ 
\multicolumn{1}{l|}{Clip-VIP~\cite{xue2022clip}} & \multicolumn{1}{c|}{ICLR'23} & \multicolumn{1}{c|}{HDV100M+W2M+C12M} & 50.5 & 78.4 & 87.1 & 72.00 &\multicolumn{1}{c|}{1} & 54.2 & 77.2 & 84.8 &72.07 & \multicolumn{1}{c|}{1} & 29.4 & 50.6 & 59.0 & 46.33 &5.0\\ 

\midrule
\multicolumn{1}{l|}{\textbf{STGT(ours)$_0$}} & \multicolumn{1}{c|}{-} & \multicolumn{1}{c|}{W2M+C3M} & 52.0 & 75.8 & 85.3 &71.03 & \multicolumn{1}{c|}{1} & 44.8 & 71.8 & 80.5 & 65.70 &\multicolumn{1}{c|}{2} & 28.2 & 51.3 & 59.6 & 46.37 & 5.0\\

\multicolumn{1}{l|}{\textbf{STGT(ours)$_1$}} & \multicolumn{1}{c|}{-} & \multicolumn{1}{c|}{W10M} & 52.2 & 76.4 & 85.0 &71.90 & \multicolumn{1}{c|}{1} & 46.9 & 73.3 & 81.7 &67.3 & \multicolumn{1}{c|}{2} & 30.3 & 52.4 & 61.9 & 48.19 &5.0\\

\multicolumn{1}{l|}{\textbf{STGT(ours)$_2$}} & \multicolumn{1}{c|}{-} & \multicolumn{1}{c|}{W10M+V4M} & 60.6 & \textbf{84.1} & \textbf{89.1} &\underline{77.93} & \multicolumn{1}{c|}{1} & \underline{53.8} & 77.9 & 84.9 &72.2 & \multicolumn{1}{c|}{1} & \underline{35.4} & \textbf{57.1} & \textbf{65.1} & \textbf{52.54} &\textbf{3.0}\\ 

\multicolumn{1}{l|}{\textbf{STGT(ours)$_3$$^\ddag$}} & \multicolumn{1}{c|}{-} & \multicolumn{1}{c|}{W10M+V4M} & 60.2 & 83.5 & 81.2 &77.30 & \multicolumn{1}{c|}{1} & 53.2 & \underline{78.5} & \textbf{86.1} &\underline{72.6} & \multicolumn{1}{c|}{1} & \textbf{35.6} & 56.8 & \underline{64.4} & \underline{52.28} &\underline{4.0} \\ 

\multicolumn{1}{l|}{\textbf{STGT(ours)$_4$}} & \multicolumn{1}{c|}{-} & \multicolumn{1}{c|}{W10M+V7M} & \textbf{61.9} & \underline{83.1} & \underline{89.0} &\textbf{78.00} & \multicolumn{1}{c|}{\textbf{1}} & \textbf{55.8} & \textbf{79.3} & \underline{85.4} &\textbf{73.5} & \multicolumn{1}{c|}{\textbf{1}} & 35.2 &56.4  &64.1  & 51.87  &\underline{4.0}\\ 

\midrule
\multicolumn{16}{c}{\textbf{Zero-shot}} \\ \midrule
\multicolumn{1}{l|}{Frozen \cite{Frozen}} & \multicolumn{1}{c|}{ICCV'21} & \multicolumn{1}{c|}{W2M+C3M}  & 21.1 & 46.0 & 56.2 &41.10 & \multicolumn{1}{c|}{7} & 18.7 & 39.6 & 51.6 &36.63 & \multicolumn{1}{c|}{10} & 9.3 & 22.0 & 30.1 & 20.47 &51.0\\
\multicolumn{1}{l|}{ALPRO \cite{ALPRO}} & \multicolumn{1}{c|}{CVPR'22} & \multicolumn{1}{c|}{W2M+C3M}  & 23.8 & 47.3 & 57.9 &43.00 & \multicolumn{1}{c|}{6} & 24.1 & 44.7 & 55.4 &41.40 & \multicolumn{1}{c|}{8} & - & - & - & - &-\\
\multicolumn{1}{l|}{OA-Trans \cite{wang2022object}} & \multicolumn{1}{c|}{CVPR'22} & \multicolumn{1}{c|}{W2M+C3M}  & 23.5 & 50.4 & 59.8 &44.57 & \multicolumn{1}{c|}{6} & 23.4 & 47.5 & 55.6 &42.17 & \multicolumn{1}{c|}{8} & - & - & - & - &-\\
\multicolumn{1}{l|}{MILES \cite{ge2022miles}} & \multicolumn{1}{c|}{ECCV'22} & \multicolumn{1}{c|}{W2M+C3M}  & 27.2 & 50.3 & 63.6 &47.03 & \multicolumn{1}{c|}{5} & 26.1 & 47.2 & 56.9 &43.30 & \multicolumn{1}{c|}{7} & 11.1 & 24.7 & 30.6 & 22.13 &50.7 \\ 
\multicolumn{1}{l|}{MCQ \cite{ge2022bridging}} & \multicolumn{1}{c|}{CVPR'22} & \multicolumn{1}{c|}{W2M+C3M}  & 25.6 & 50.6 & 61.1 &25.77 &  \multicolumn{1}{c|}{5} & 26.0 & 46.4 & 56.4 &42.93 & \multicolumn{1}{c|}{7} & 12.2 & 25.9 & 32.2 & 23.43 &42.0\\ 
\multicolumn{1}{l|}{CLIP4Clip~\cite{clip4clip}} & \multicolumn{1}{c|}{ NeurCom'22} & \multicolumn{1}{c|}{W10M+H100M} & - & - & - &- & \multicolumn{1}{c|}{-} & 31.2 & 53.7 & 64.2 &49.70 & \multicolumn{1}{c|}{4} & - & - & - & - \\ 
\multicolumn{1}{l|}{Clover~\cite{huang2023clover}} & \multicolumn{1}{c|}{CVPR'23} & \multicolumn{1}{c|}{W2M+C3M}  & 29.5 & 55.2 & 66.3 &50.33 & \multicolumn{1}{c|}{4} & 26.4 & 49.5 & 60.0 &45.30 & \multicolumn{1}{c|}{6} & 14.7 & 29.2 & 38.2 & 27.37 &\underline{24.0}\\
\multicolumn{1}{l|}{GLSCL~\cite{tu2023global}} & \multicolumn{1}{c|}{CVPR'23} & \multicolumn{1}{c|}{W2M+C3M+VG+SBU}  & - & - & - &- & \multicolumn{1}{c|}{-} & 30.2 & 52.3 & 62.7 &48.40 & \multicolumn{1}{c|}{-} & 17.3 & 33.0 & 39.2 & 29.83 &-\\
\multicolumn{1}{l|}{Imagebind~\cite{girdhar2023imagebind}} & \multicolumn{1}{c|}{CVPR’23} & \multicolumn{1}{c|}{\textgreater 100M}  & -& - & - &- & \multicolumn{1}{c|}{\textbf{-}} & 36.8 & 61.8 & 70 &56.20 & \multicolumn{1}{c|}{-} & - & - & - & - &-\\

\midrule

\multicolumn{1}{l|}{\textbf{STGT(ours)$_0$}} & \multicolumn{1}{c|}{-} & \multicolumn{1}{c|}{W2M+C3M}  & 34.0 & 61.7 & 68.1 &54.6 & \multicolumn{1}{c|}{3} & 32.8 & 52.5 & 60.2 & 48.50 &\multicolumn{1}{c|}{5} & 15.4 & 29.6 & 37.5 & 27.50 & 31.0\\ 

\multicolumn{1}{l|}{\textbf{STGT(ours)$_1$}} & \multicolumn{1}{c|}{-} & \multicolumn{1}{c|}{W10M}  & 34.7 & 61.2 & 68.7 &54.87 & \multicolumn{1}{c|}{3}& 34.4 & 55.6 & 63.8  &51.27 &\multicolumn{1}{c|}{4} & 16.0 & 30.7 & 38.7 & 28.47 &31.0\\ 

\multicolumn{1}{l|}{\textbf{STGT(ours)$_2$}} & \multicolumn{1}{c|}{-} & \multicolumn{1}{c|}{W10M+VIDAL4M}  & \textbf{39.9} & \textbf{64.5} & \textbf{72.0} &\textbf{58.79} & \multicolumn{1}{c|}{3} & {39.8} & {63.5} & {73.1} &58.80 &\multicolumn{1}{c|}{3} & \underline{21.1} & \underline{35.0} & \underline{41.7} & \underline{32.64} &\underline{24.0}\\

\multicolumn{1}{l|}{\textbf{STGT(ours)$_3$$^\ddag$}} & \multicolumn{1}{c|}{-} & \multicolumn{1}{c|}{W10M+VIDAL4M}  & \textbf{40.0} & \underline{61.8} & \underline{70.9} &\underline{57.56} & \multicolumn{1}{c|}{3} & \underline{40.0} & \underline{63.7} & \underline{73.3} &\underline{59.00} & \multicolumn{1}{c|}{3} & 19.5 & 33.7 & 40.5 & 31.23 &24.5\\

\multicolumn{1}{l|}{\textbf{STGT(ours)$_4$}} & \multicolumn{1}{c|}{-} & \multicolumn{1}{c|}{W10M+VIDAL7M}  & 37.8 & 61.5 & 70.3 &56.53 & \multicolumn{1}{c|}{3} & \textbf{40.9} & \textbf{64.0} & \textbf{73.8} &\textbf{59.57} & \multicolumn{1}{c|}{3} & \textbf{20.6} & \textbf{37.5} & \textbf{44.5} & \textbf{34.20} &\textbf{21.5}\\ 
\bottomrule
\end{tabular}}
\vspace{-1.5em}
\label{tab:retrieval_SOTA}
\end{table*}

\subsection{Implementation Details}
We initialize the video encoder with CLIP (ViT-g)~\cite{CLIP}, and the text encoder with ~\cite{blip2}. Our pre-training is divided into two stages. In the first stage, we use contrastive loss with an initial learning rate of 5$e$-5, the batch size is set to 20 on a single GPU, and train 12 epochs. In the second stage, we adopt cross-similarity alignment loss with an initial learning rate of 2.5$e$-5, the batch size is set to 20, and train for 6 epochs. The pre-training is performed entirely on 48 A100 GPUs. We use the AdamW optimizer with a weight decay 0.05 and betas (0.9, 0.98). All video frames are resized to 224 $\times$ 224, each frame is split into patches with a size of 16 $\times$ 16, and we randomly sample $m=8$ frames per video. For retrieval tasks, we only finetune
for 9 epochs with the cross-similarity alignment loss and an initial learning rate of 2.5$e$-5 on 8 A100 GPUs. For fine-tuning on video QA, we adopt the BLIP-2~\cite{blip2} and add a large language model (OPT-2.7B and Vicuna-7B) as the dialogue VQA system after Q-Former. At this stage, we fine-tune with the answer generation loss to only train some parameters of Q-Former and fully connected layer as~\cite{blip2} to align vision encoding with LLM with an initial learning rate 3$e$-5 for 9 epochs. By default, video and text embeddings are 256 dimensions.

\begin{table}[tp]
        \renewcommand{\arraystretch}{0.7}
	\begin{center}
     \caption{Ablation study for topology ($\mathcal{A}$) and weights ($\mathcal{W}s $) of spatio-temporal graphs Transformer. } \label{tab:STRGT}
     \vspace{-1.0em}
        \scalebox{0.85}{
    \begin{tabular}{lcccccccc}
        \toprule 
        w/o  $\mathcal{A}$      &  w/o $\mathcal{W}s $  & R@1$\uparrow$   & R@5$\uparrow$ & R@10$\uparrow$ &R-Mean$\uparrow$  & $\Delta$ \\ \midrule
        \midrule
         Baseline & - & 30.1 & 49.3 & 57.3 & 45.57 & -\\
        \XSolidBrush & \XSolidBrush & 30.0 & 51.1 & 59.1 & 46.73 & +1.16\%\\
         \CheckmarkBold  & \XSolidBrush  & 32.4 & 52.8 & 61.4 & 48.86 &+3.29\% \\
                 \CheckmarkBold & \CheckmarkBold  & \textbf{33.1} & \textbf{54.2} & \textbf{61.7} & \textbf{49.67} &\textbf{+4.10\%} \\
        \bottomrule
    \end{tabular}
    }
	\end{center}
 \vspace{-2.0em}
\end{table}

\subsection{Ablation Study}
We conduct ablation experiments on MSRVTT to evaluate the effects of  $\hbar$ and $\gamma$  in $\mathcal{L}_{csal}$, as well as the STGT module and $\mathcal{L}_{csal}$. These experiments focus on zero-shot text-to-video retrieval, only Webvid-10M for pre-training.

\textbf{Similarity Threshold $\hbar$.}  During pre-training, we set a default constant value of  $\hbar$ ($\hbar = 0.1$) to ensure that the graph edges aren't too sparse, thereby enabling the STGT to effectively learn how to dynamically sparsify the edges in spatio-temporal graph. In testing, we verify the impact of  $\hbar$ on zero-shot text-to-video retrieval on MSRVTT to select an appropriate value. As shown in Fig.~\ref{fig:hyperparameter}~(a), our STGT outperforms the `baseline' when  $\hbar \geq 0$ with contrastive loss. The performance (R-Mean) gradually increases and then decreases as 
$\hbar$ increases in the range of 0 to 1.0. The peak performance of approximately 49.67\% is achieved with a similarity threshold of $\hbar = 0.5$, surpassing the `baseline' (dashed green line) by 4.1\% in terms of R-Mean.

\textbf{Temperature Coefficient $\gamma$.}  After the initial training with contrastive learning loss, we use the cross-similarity alignment loss ($\mathcal{L}_{csal}$) to further optimize the model. The choice of $\gamma$ significantly influences the performance of zero-shot text-to-video retrieval on MSRVTT. Hence, we conduct experiments with integer values of 
$\gamma$ ranging from 1 to 12. As shown in Fig.~\ref{fig:hyperparameter}~(b), the performance (R-Mean) initially increases and then decreases as 
$\gamma$ increases. When  $\gamma$ is less than 3, $\mathcal{L}_{csal}$ provides minimal gains. However, when $\gamma$ is between 3 and 10, $\mathcal{L}_{csal}$ contributes to a notable improvement in performance. Once $\gamma$ exceeds 10, the performance improvement becomes negligible because the $\mathcal{L}_{csal}$ and contrast loss ($\mathcal{L}_{vtc}$) become almost equivalent in this case.

\begin{table}[htbp]
         \renewcommand{\arraystretch}{0.7}
        \caption{Ablation study of STGT and $\mathcal{L}_{csal}$.} \label{tab:STGT_lssa}
        \vspace{-1.5em}
	\begin{center}
		\label{tbablation_center_region}
            \scalebox{0.85}{
		\begin{tabular}{lcccccccc}
					\toprule 
					Methods        & STGT  & $\mathcal{L}_{csal}$ & R@1$\uparrow$   & R@5$\uparrow$ & R@10$\uparrow$ &R-mean$\uparrow$ \\ \midrule
					\midrule
					Baseline					& \XSolidBrush & \XSolidBrush & 30.1 & 49.3 & 57.3 & 45.57 \\
					\textbf{Ours Method}				& \CheckmarkBold  & \XSolidBrush  & 33.1 & 54.2 & 61.7 & 49.67  \\
    			\textbf{Ours Method} & \CheckmarkBold & \CheckmarkBold  & \textbf{34.4} & \textbf{55.6} & \textbf{63.8} & \textbf{51.27}  \\
					\bottomrule
				\end{tabular}
    }
   \end{center}
 \vspace{-1.2em}
\end{table}

\textbf{STGT and $\mathcal{L}_{csal}$.}  As listed in Tab.~\ref{tab:STRGT}, the results demonstrate that the spatio-temporal graph ($\mathcal{A}$) brings a significant performance improvement of 3.29\% in R-Mean. Additionally, the weight information ($\mathcal{W}s $) brings  a slight improvement of 0.71\%. Overall, the STGT achieves an impressive 4.1\% improvement in R-Mean. Furthermore, cross-similarity alignment loss ($\mathcal{L}_{csal}$) brings a considerable 1.6\% improvement in R-Mean, as well as comprehensive improvement across the board for R@1, R@5, and R@10 metrics.

\begin{table}[!t]
    \renewcommand{\arraystretch}{0.7}
    \setlength{\tabcolsep}{2pt}
    \centering
    \caption{Results of text-to-video retrieval on MSVD dataset.}
    \label{tab:result_of_retrieval_msvd}
    \vspace{-0.5em}
    \scalebox{0.95}{
        \begin{tabular}{lcccccc}
            \toprule
            Methods  & R@1$\uparrow$   & R@5$\uparrow$ & R@10$\uparrow$ & R-mean $\uparrow$  & MedR$\downarrow$ \\ \midrule
            CE~\cite{radford2021learning}   & 19.8 & 49.0 & 63.8 &44.2 &6.0 \\ 
            SUPPORT~\cite{patrick2020support} & 28.4 & 60.0 & 72.9 &53.77 & 4.0 \\ 
            Frozen~\cite{Frozen}	& 33.7 & 64.7 & 76.3 &58.23 & 3.0 \\
            CLIP4Clip~\cite{clip4clip}& 46.2 & 76.1 & 84.6 &68.97 &2.0 \\
            \midrule
            \textbf{STGT(ours)$_1$} & 52.1 & 80.6 &87.4 &73.37&\textbf{1.0} \\
            \textbf{STGT(ours)$_2$} & 55.6 & 81.9 &\underline{88.2} &75.21&\textbf{1.0} \\
            \textbf{STGT(ours)$_3$$^\ddag$} & \underline{56.2} & \underline{82.}&\textbf{88.8} &\underline{75.90} &\textbf{1.0} \\
            \textbf{STGT(ours)$_4$} & \textbf{56.3} & \textbf{82.8} &\textbf{88.8} &\textbf{75.95} & \textbf{1.0}\\
            \bottomrule
        \end{tabular}
    }
    \vspace{-1.8em}
\end{table}

\begin{figure*}[tp]
\setlength{\abovecaptionskip}{-0.00cm}
    \begin{center}
    \includegraphics[width=1.0\linewidth]{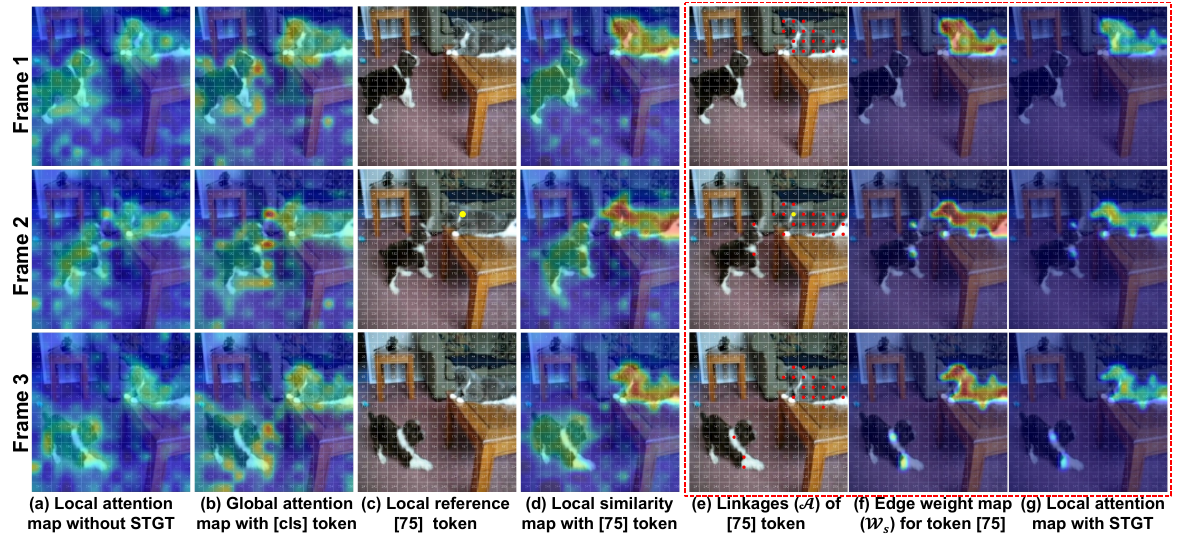}
    \caption{The visualizations of a reference token on MSVD, including (a) local attention map without STGT, (b) Global attention map with $[cls]$ token, (c) local reference token located in 75, (d) local similarity map with reference token, (e) link relationship $\mathcal{A}$ of reference token with $\hbar = 0.5$, (f) weight of edge in $\mathcal{W}_s$, (g) local attention map with our STGT. }
    \label{fig:exp-vis1}
    \end{center}
    \vspace{-1.6em}
\end{figure*} 

\vspace{-1.0em}
\subsection{Comparing to State-of-the-art}

\textbf{Text-to-video retrieval.} Tab.~\ref{tab:retrieval_SOTA}  and Tab.~\ref{tab:result_of_retrieval_msvd} show the retrieval results on the DiDeMo, MSRVTT, LSMDC, and MSVD datasets with both zero-shot and fine-tuning settings. Our STGT significantly outperforms previous works across all datasets, demonstrating its superior generalization ability, particularly under zero-shot evaluation. Specifically, in the zero-shot text-to-video retrieval task, our method achieves a 13.7\% gain in R-Mean over Clover, a 9.3\% gain in R-Mean over CLIP4Clip, and a 2.8\% gain in R-Mean over Imagebind on the MSRVTT. Compared to previous state-of-the-art works on challenging LSMDC dataset, our method surpasses GLSCL by 4.37\% in R-Mean and outperforms Clover by 6.83\% in R-Mean. Our STGT also excels in all R@1, R@5, and R@10 metrics.
When fine-tuned on downstream datasets, our STGT still demonstrates superior performance, outperforming previous state-of-the-art methods. Our method achieves remarkable performance on four datasets, surpassing previous works by a significant margin across most metrics. The extensive experiments show that the performance improvements brought about by data augmentation and model enhancement are substantial and should not be overlooked.

\begin{table}[tp]
\renewcommand{\arraystretch}{0.65}
\caption{Comparisons with existing methods on \textbf{MSRVTT-QA} and \textbf{MSVD-QA} in top-1 accuracy (\%). JuskAsk~\cite{yang2021just} uses 69M QA domain-specific data to pre-train their model.}\label{tbl:qa}
\vspace{-1.2em}
\resizebox{0.45\textwidth}{!}{
    \aboverulesep = 0.55mm
    \belowrulesep = 0.55mm
	\small
	\centering	
        \begin{tabular}	{l  c |  c c }
	\toprule
	\textbf{Method} & \textbf{Pre-training dataset} & \textbf{MSRVTT}     & \textbf{MSVD} \\
	\midrule
	ClipBERT~\cite{Clipbert} & COCO + VG (5.6M) & 37.4 & - \\
        SSML~\cite{amrani2021noise} & H100M & 35.1 & 35.1 \\
	CoMVT~\cite{seo2021look} & H100M & 39.5 & 42.6 \\
        JuskAsk~\cite{yang2021just} & HTVQA69M & 41.5 & 46.3 \\
        MERLOT~\cite{zellers2021merlot} & Y180M & 43.1 & - \\
        VIOLET~\cite{fu2021violet} & W2M+C3M+Y180M & 43.9 & 47.9 \\
        ALPRO~\cite{ALPRO} & W2M+C3M & 42.1 & 45.9 \\
        All-in-1~\cite{wang2023all} & W2M+H100M & 42.9 & 47.9 \\
        Clover~\cite{huang2023clover} & W2M+C3M & 44.1 & 52.4 \\
	\midrule
        \textbf{STGT+OPT$_{2.7B}$} & W10M+V7M & \underline{51.2} & \underline{60.3} \\
        \textbf{STGT+Vicuna$_{7B}$} & W10M+V7M & \textbf{58.8} & \textbf{69.7} \\
        \bottomrule
	\end{tabular}
    }
\vspace{-1.8em}
\end{table}
 
\textbf{Video question answering.}  Tab.~\ref{tbl:qa} shows the comparative results of our method and existing state-of-the-art (SOTA) methods on two video QA datasets. The models STGT+OPT$_{2.7B}$ and STGT+Vicuna$_{7B}$ outperform other methods on both datasets, demonstrating our method's superior performance. Notably, when compared with the current SOTA method, Clover~\cite{huang2023clover}, our STGT+OPT$_{2.7B}$ model achieves  significant improvements of 7.1\% and 7.9\% on the MSRVTT-QA and MSVD-QA, respectively. Similarly, STGT+Vicuna$_{7B}$ shows impressive improvements of 14.7\% and 17.3\% on the MSRVTT-QA and MSVD-QA, respectively. These experimental results further highlight the superiority and effectiveness of our method.

\begin{table}[htbp]
\renewcommand{\arraystretch}{0.6}
\begin{center}
    \caption{comparison between parameters and efficiency. We calculated FPS for processing 1K videos on the MSRVTT testset} \label{tab:efficiency}
     \vspace{-1.0em}
    \scalebox{0.75}{
    \begin{tabular}{lcccccccc}
        \toprule 
        method &  Spatio–Temporal parameters & Learnable parameters   &total time (s) & FPS \\ \midrule
        \midrule
         ALPRO & - & 230.5M &472 & 2.12 \\
         Baseline & 60.4M (CCT+MIT)~\cite{X-CLIP}  & 247.1M  &276& 3.62 \\
         \textbf{STGT(ours)} & 59.5M (STGT) & 246.2M &252& 3.97 \\
         \bottomrule
    \end{tabular}
}
\end{center}
 \vspace{-2.0em}
\end{table}

\textbf{Model parameter and efficiency analysis.} We perform comparative experiments for our method and ALPRO~\cite{ALPRO} on 8 A100 GPUs. As listed in Tab.~\ref{tab:efficiency}, despite a slight increase in parameters by 16M, our method nearly doubles ALPRO's efficiency due to the sparse design of spatio-temporal graphs transformer, showcasing its superior performance. In other words, our STGT module is straightforward and does not significantly increase complexity.

\begin{figure*}[htbp]
\setlength{\abovecaptionskip}{-0.00cm}
    \begin{center}
    \includegraphics[width=0.95\linewidth]{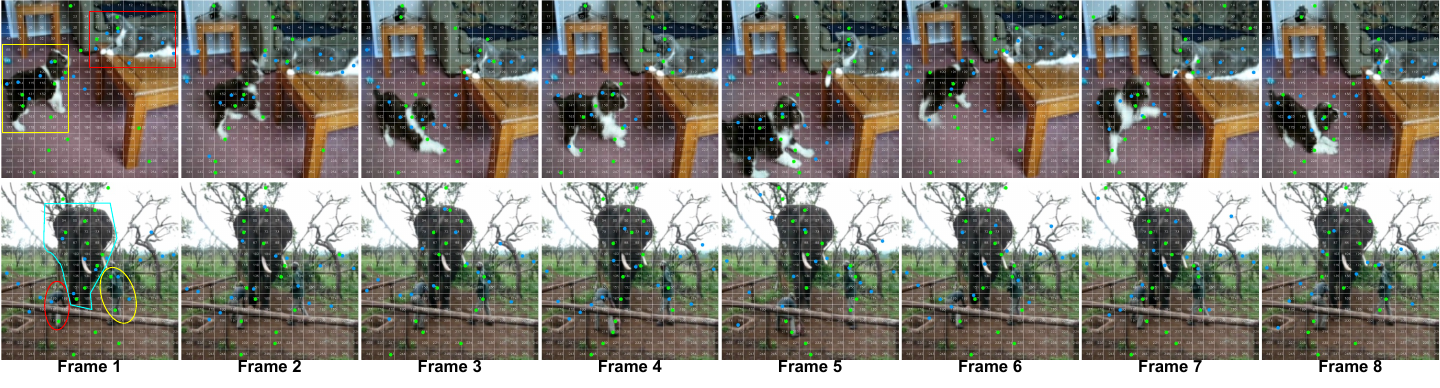}
    \caption{Visualization of max-pooling sampling position in local tokens. The blue point represent the sampling point positions in the $x$-axis, and the green point represent the sampling point positions in the $y$-axis. }
    \label{fig:exps_maxpooling} 
    \end{center}
    \vspace{-1.8em}
\end{figure*} 

\begin{figure}[tp]
\setlength{\abovecaptionskip}{-0.00em}
	\begin{minipage}[t]{0.96\linewidth}
		\includegraphics[width=0.96\linewidth]{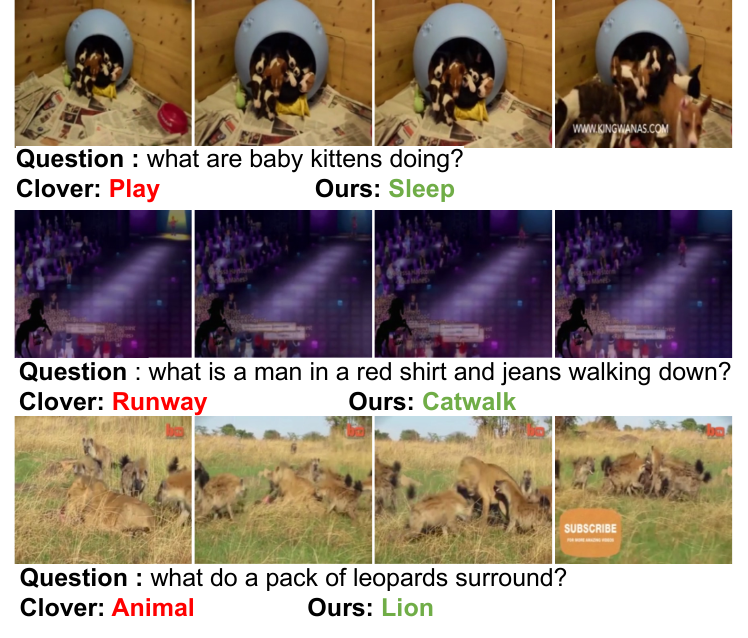}
		\caption{ Qualitative results on MSRVTT-QA.}
	\label{fig:exp_msrvtt_qa}
 \end{minipage}%
 \vspace{-2.0em}
\end{figure}

\subsection{Visual Analysis of Experimental Results}

\textbf{Visual mas of reference token.}  In Fig.~\ref{fig:exp-vis1},  we provide several visualizations of a reference token to illustrate the influence of the proposed spatio-temporal graph transformer.  As shown in Fig.~\ref{fig:exp-vis1} (a) and (b), when spatio-temporal graph is not employed in attention, the attention map (Fig.~\ref{fig:exp-vis1} (a)) with the reference token is slightly similar to the global attention map with  $[cls]$ token (Fig.~\ref{fig:exp-vis1} (b)). This indicates a trend of assimilation between local and global tokens, with local tokens not specifically focusing on the local features of the same object. To encourage local tokens to pay more attention to the same object, we designed an innovative spatio-temporal graph transformer module. This module uses spatio-temporal graphs to constrain local attention, and allow local and global tokens to each play their distinct roles, thereby enhancing local features.  As shown in Fig.~\ref{fig:exp-vis1} (e) and (f), for reference token $[74]$ marked with a yellow point, we establish spatial and temporal linkage relationships using its local similarity within frames or between adjacent frames (tokens marked with a red point indicate a linking relationship with the reference tokens). These similarities are also used to assign weights for edges in the spatio-temporal graph.
The topology of the graph and the weights of the edges are incorporated into the attention mechanism by Eq.~\ref{EQ_Mask} and Eq.~\ref{EQ_Atten}. As a result, the attention of local reference token pay more attention to local object (such as cat), and local attention can also track the same object (such as cat) well between different frames.

\textbf{Max-Pooling Sampling.} In Fig.~\ref{fig:exps_maxpooling}, we visualize the the location of max-pooling sampling point in the $x$-axis and $y$-axis directions. Given the local token's feature dimension of $16\times16\times1408$, we have $16\times1408$ sampling points along the $x$-axis and $y$-axis. For simplicity, we average the positions of these sampling points within the feature dimension. Consequently, in Fig.~\ref{fig:exps_maxpooling}, each frame image has 16 sampling points on $x$-axis and $y$-axis, respectively. As shown in Fig.~\ref{fig:exps_maxpooling}, the sampling points are densely clustered in the primary object region and sparsely distributed in background, as highlighted by the regions enclosed by yellow, red, and blue lines. This indicates that these sampling points effectively capture the features of these local regions. When combined with global features, they facilitate video understanding while minimizing computational complexity.

\begin{figure}[tp]
\vspace{-0.5em}
\setlength{\abovecaptionskip}{-0.00em}
	\begin{minipage}[t]{0.96\linewidth}
		\includegraphics[width=0.96\linewidth]{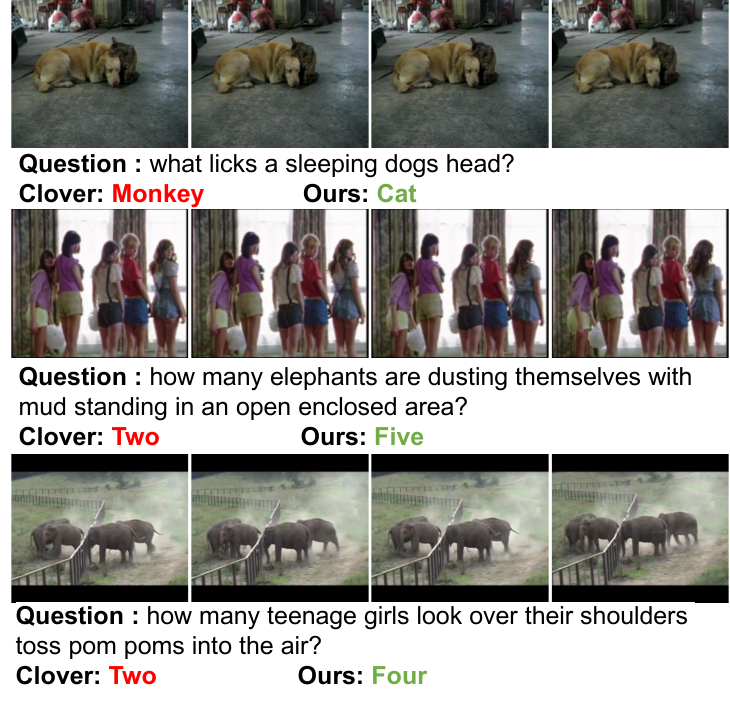}
		\caption{Qualitative results on MSVD-QA.} 
        \label{fig:exp_msvd_qa}
        \end{minipage}%
\vspace{-2.0em}
\end{figure}

\begin{figure}[htbp]
\vspace{-1.2em}
\setlength{\abovecaptionskip}{-0.2em}
\centering
\subfigure[T2V Retrieval on DiDeMo]{
    \begin{minipage}[t]{0.49\linewidth}
	\includegraphics[width=4.05cm,height=3cm]{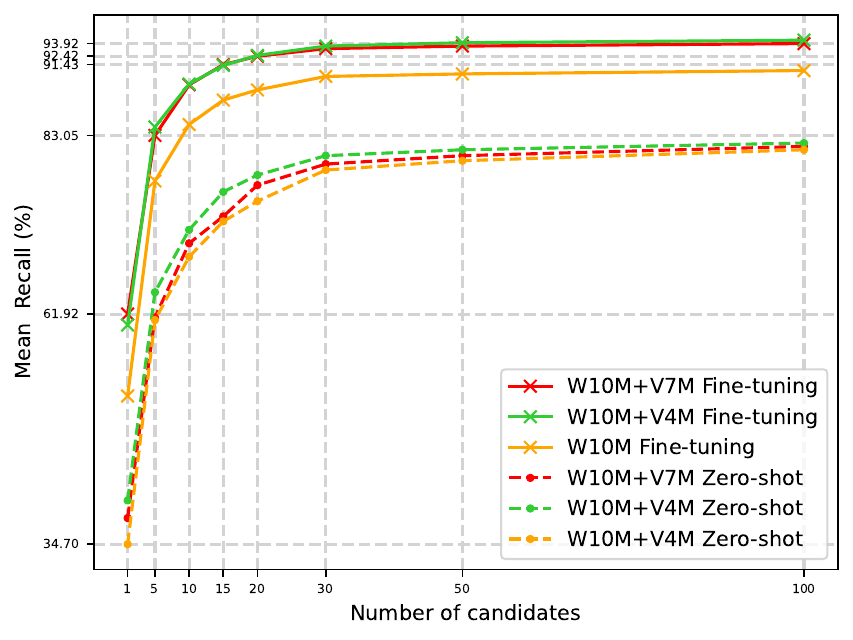}
    \end{minipage}
}%
\subfigure[V2T Retrieval on DiDeMo]{
    \begin{minipage}[t]{0.49\linewidth}
	\includegraphics[width=4.05cm,height=3cm]{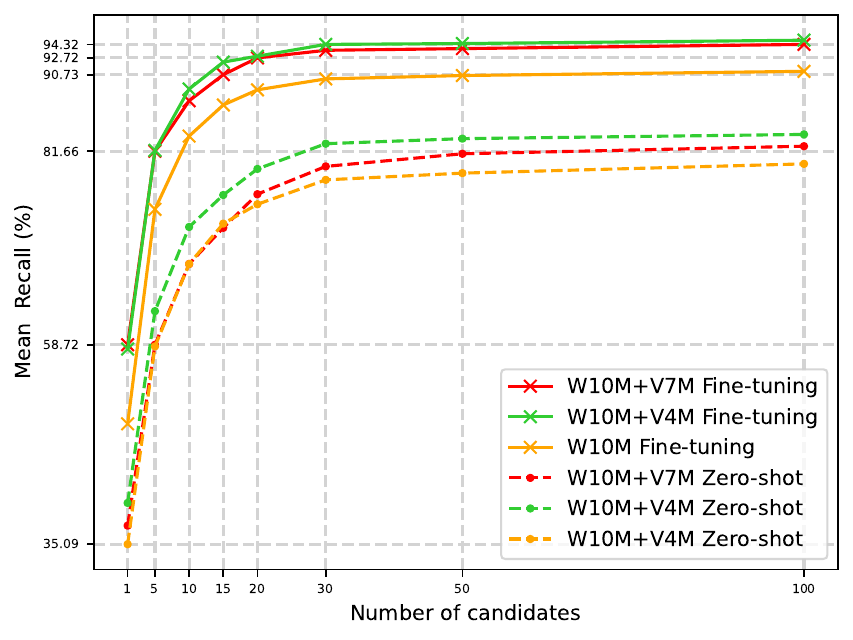}	
    \end{minipage}
}%
\quad
\subfigure[T2V Retrieval on MSRVTT]{
    \begin{minipage}[t]{0.49\linewidth}
	\includegraphics[width=4.05cm,height=3.3cm]{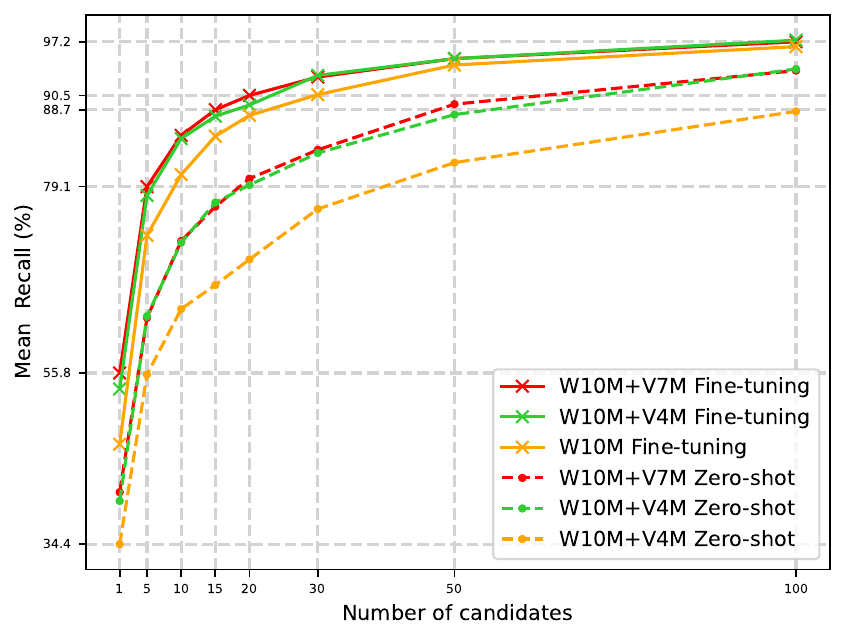}
    \end{minipage}
}%
\subfigure[V2T Retrieval on MSRVTT]{
    \begin{minipage}[t]{0.49\linewidth}
        \includegraphics[width=4.05cm,height=3.3cm]{imgs/didemo_tr.pdf}		
    \end{minipage}
}%
\caption{The retrieval accuracy versus the number of candidates for fine-tuning and zero-shot with different pre-taining data on DideMo and MSRVTT.}
\label{fig:fig_recall_carves}
\centering
\vspace{-1.6em}
\end{figure}

\begin{figure*}[tp]
    \setlength{\abovecaptionskip}{-0.3cm}
    \begin{center}
    \includegraphics[width=0.96\linewidth]{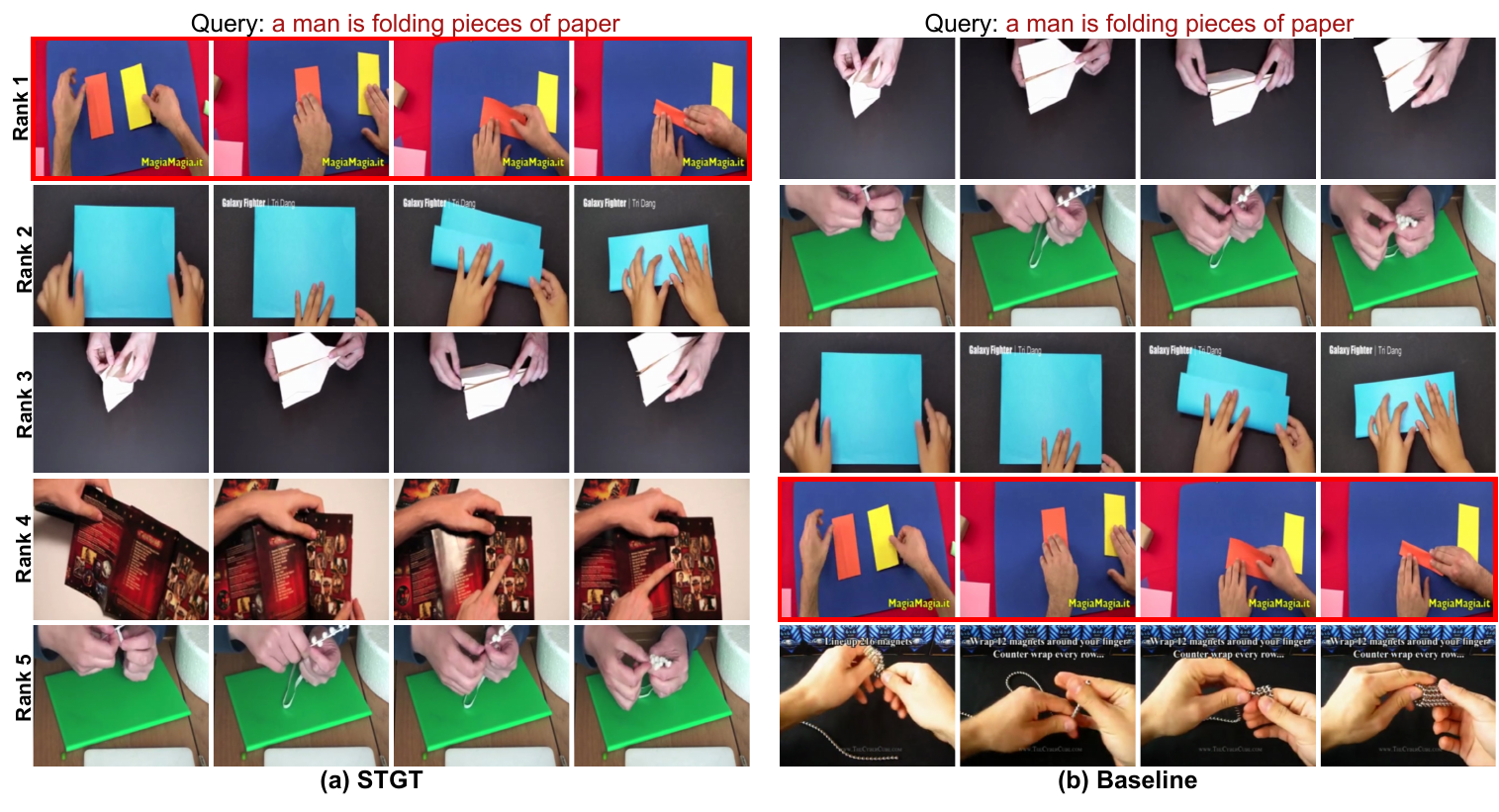}
    \caption{Qualitative results of  text-to-video retrieval results on MSRVTT. }
    \label{fig:exp_vis_t2v}
    \end{center}
    \vspace{-1.6em}
\end{figure*} 

\begin{figure*}[tp]
\setlength{\abovecaptionskip}{-0.3cm}
    \begin{center}
    \includegraphics[width=0.98\linewidth]{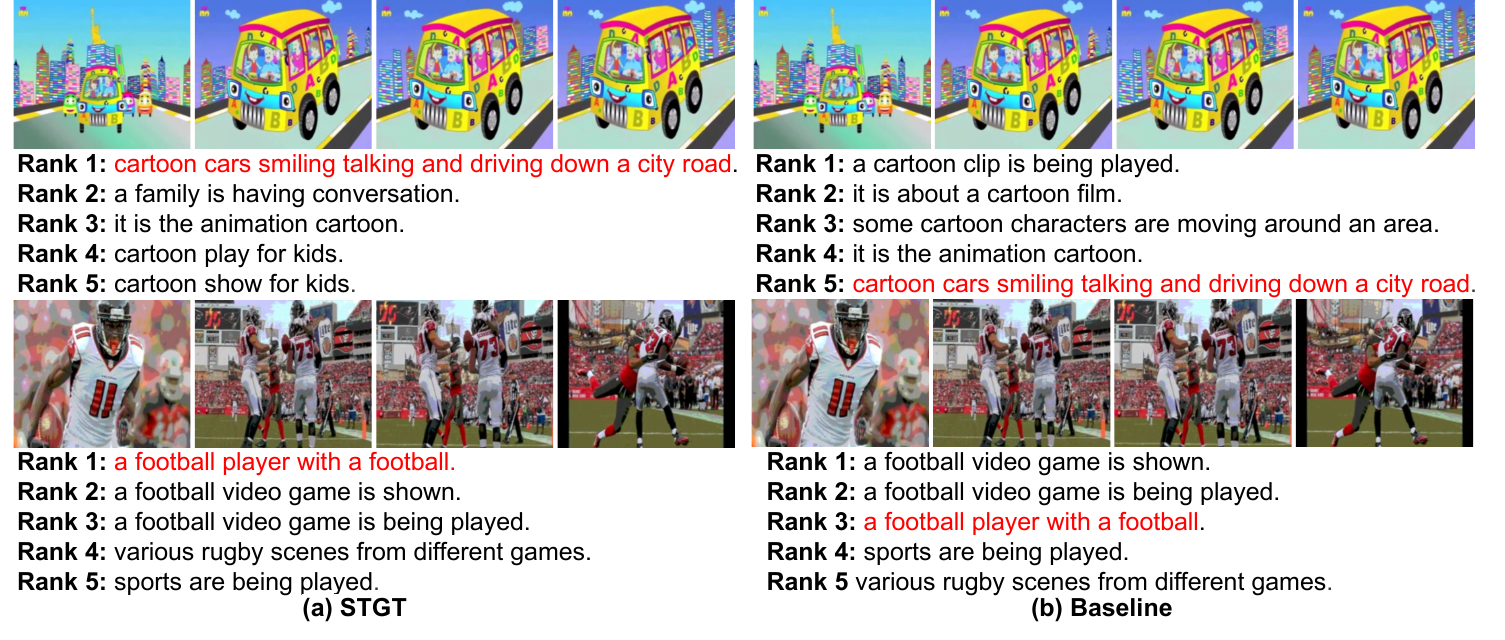}
    \caption{Qualitative results of video-to-text retrieval results on MSRVTT. }
    \label{fig:exp_vis_v2t}
    \end{center}
    \vspace{-2.0em}
\end{figure*} 

\textbf{Video Question Answering.} We evaluate our dialog model with Vicuna-7B on MSR-VTT-QA and MSVD-QA datasets against the Clover. Fig.~\ref{fig:exp_msrvtt_qa} shows our STGT higher accuracy in video understanding and question answering. For example, it correctly answers `Sleep' for what baby kittens are doing, while Clover says `Play'. Our STGT also identifies `Catwalk' for a man walking down, compared to Clover's `Runway', and accurately recognizes a `Lion' instead of a general `Animal'. Fig.~\ref{fig:exp_msvd_qa} highlights our STGT superiority. It identifies a `Cat' licking a dog's head, not a `Monkey' as Clover responses. It counts `Four' elephants, more accurate than Clover's `Two'. The proposed STGT correctly answers `Five' girls, versus Clover's `Two'. Overall, our  method outperforms in detail recognition and accuracy, indicating its potential in video comprehension, and providing insights for future advancements.

\begin{figure*}[htbp]
\setlength{\abovecaptionskip}{-0.10cm}
    \begin{center}
    \includegraphics[width=0.99\linewidth]{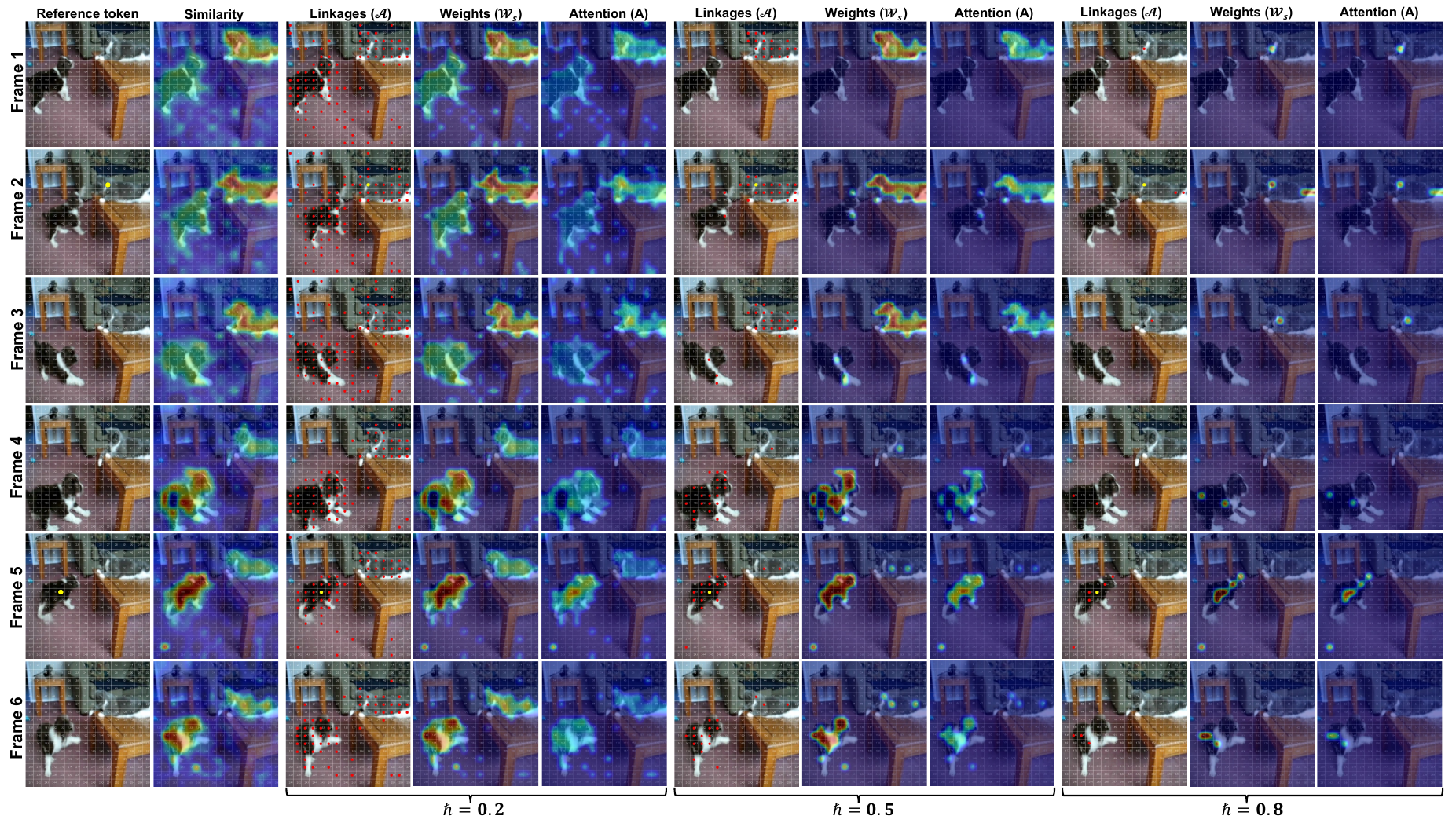}
    \caption{The linkages and attention distribution of reference tokens (cat and dog) with different $\hbar$. }
    \label{fig:exps_sm1}
    \end{center}
    \vspace{-1.5em}
\end{figure*}

\textbf{Retrieval Accuracy Curves.} Fig.~\ref{fig:fig_recall_carves} provides several curves of the retrieval accuracy versus the number of candidates on DiDeMo and MSRVTT datasets. Notably,  beyond a certain threshold, the impact of increasing the number of candidates on the final results becomes negligible. For fine-tuned models, the retrieval accuracy essentially plateaus when the number of candidate samples reaches 30. It indicates that adding more candidates beyond this point does not significantly improve the model's performance. For zero-shot models, this threshold is higher, indicating that these models may benefit from a larger number of candidates. However, it's important to note that even with a higher threshold, the peak detection accuracy of zero-shot models does not surpass that of the fine-tuned models. This observation highlights the effectiveness of fine-tuning in improving model performance, even when the number of candidate samples is limited.

\textbf{Retrieval Comparison.} In Fig.~\ref{fig:exp_vis_t2v} and Fig.~\ref{fig:exp_vis_v2t}, we present a visual comparison of the retrieval results between our spatio-temporal graph transformer (STGT) and the 'Baseline' model. In Fig.~\ref{fig:exp_vis_t2v}, the 'Baseline' model, influenced by the visual similarity of videos, identifies three origami videos (Rank 1, Rank 3, and Rank 4) and two handicraft videos (Rank 2 and Rank 5). However, the correct video is only placed at Rank 4. In contrast, our model, enhanced with STGT and cross-similarity alignment loss, accurately identifies four origami videos (Rank 1~4) and one handicraft video (Rank 5), with the correct video securing the top rank. This clearly illustrates the superior accuracy of our model. 

In Fig.~\ref{fig:exp_vis_v2t}, we offer some visualized results for video-to-text retrieval. It's clear from Fig.~\ref{fig:exp_vis_v2t} that our model excels not only in accurately retrieving the correct text using video embedding, but also in retrieving a variety of other texts that correspond with the video scene. Even though these texts are not the standard answers in the test dataset, they appear to be apt descriptions of the video content. This indicates that our model possesses a more comprehensive understanding and can generate diverse, yet contextually relevant, descriptions for the same video content.

\textbf{Effect of hyperparameter $\hbar$ on Spatio-Temporal Graph and Local Attention.} 
Fig.~\ref{fig:exps_sm1} provides a visual representation of the attention distribution for reference tokens under various hyperparameter ($\hbar$) settings. In Fig.~\ref{fig:exps_sm1}, the model's capacity to differentiate between `cat' and `dog' reference tokens is displayed across a range of $\hbar$ values. At lower $\hbar$ values, the attention is more dispersed, indicating a wider, albeit less precise, focus on the relevant subjects within the image. As $\hbar$ increases, the attention distribution becomes more focused, suggesting a refined concentration that enhances the model's ability to accurately distinguish between the cat and dog within the scene. However, it's important to note that if $\hbar$ is excessively large, the reference local token may not effectively capture crucial local information between adjacent frames. This could result in under utilization of the temporal information between video frames, potentially leading to a decrease in the model's feature table modeling ability. These figures collectively highlight the delicate balance required when setting the value of $\hbar$ to optimize the model's performance in identifying and focusing on specific subjects within complex visual scenes.
\section{Conclusion} \label{Conclusion}
This paper introduces a novel spatio-temporal graph transformer (STGT) module for video-language alignment. By combining the strengths of both graph and transformer, the STGT effectively learns the spatial and temporal features of videos, utilizing the spatio-temporal contexts provided by the spatio-temporal graph. Additionally, we propose a novel cross-similarity alignment loss (CSAL) to explore the inherent self-similarity via evaluating the corresponding two video-video and text-text pairs, further promoting the accuracy of video-text alignment. Experimental results on challenging downstream tasks demonstrate that our method surpasses existing state-of-the-art methods, achieving impressive performance in video-text retrieval and video question answering tasks.

\ifCLASSOPTIONcaptionsoff
  \newpage
\fi

\bibliographystyle{IEEEtran}
\bibliography{IEEEbib}

\end{document}